\pdfoutput=1

\documentclass[11pt]{article}

\usepackage[final]{acl}

\usepackage{times}
\usepackage{latexsym}

\usepackage[T1]{fontenc}

\usepackage[utf8]{inputenc}

\usepackage{microtype}

\usepackage{inconsolata}

\usepackage{graphicx}
\usepackage{times}
\usepackage{latexsym}
\usepackage[most]{tcolorbox}   
\usepackage{listings}          
\usepackage{xcolor}    
\lstdefinestyle{plain}{
    basicstyle=\fontsize{7}{9.5}\ttfamily,
    keywordstyle=\color{blue},
    commentstyle=\color{gray},
    stringstyle=\color{green},
    showstringspaces=false,
    breaklines=true,
    breakatwhitespace=false,
    breakindent=0pt,
    escapeinside={(*@}{@*)}
}
\usepackage{algorithm}
\usepackage{algpseudocode}
\usepackage{hyperref}
\usepackage{url}
\usepackage{amsmath}
\usepackage{graphicx}
\usepackage{booktabs}
\usepackage{multirow}
\usepackage{soul}
\usepackage{svg}
\usepackage{xcolor}
\usepackage{amssymb}
\usepackage{epstopdf}
\usepackage{subfigure}
\usepackage{verbatim}
\usepackage{bm}
\usepackage{array}
\usepackage{makecell}
\usepackage{xspace}
\usepackage{framed}
\usepackage{changepage}
\usepackage{cleveref}
\usepackage{enumitem}
\usepackage{siunitx}
\usepackage{tablefootnote}
\usepackage{wrapfig}

\definecolor{deepred}{rgb}{0, 0, 0}
\definecolor{deepgreen}{rgb}{0, 0, 0}
\definecolor{deepgreen_1}{rgb}{0.219, 0.462, 0.113}

\newcommand{\bench}{\textsc{RULEARN}\xspace}
\newcommand{\method}{\textsc{IDEA}\xspace}

\title{IDEA: Enhancing the Rule Learning Ability of Large Language Model Agent through Induction, Deduction, and Abduction}

\author{Kaiyu He, 
Mian Zhang, 
Shuo Yan, 
Peilin Wu,
Zhiyu Zoey Chen \\
Department of Computer Science\\
University of Texas at Dallas\\
\texttt{\{kaiyu.he, zhiyu.chen2\}@utdallas.edu}
}

\begin{document}
\maketitle
\begin{abstract}
While large language models (LLMs) have been thoroughly evaluated for deductive and inductive reasoning, their proficiency in holistic rule learning in interactive environments remains less explored. We introduce \bench, a novel benchmark to assess the rule-learning abilities of LLM agents in interactive settings. In \bench, agents strategically interact with simulated environments to gather observations, discern patterns, and solve complex problems. To enhance the rule-learning capabilities for LLM agents, we propose \method, a novel reasoning framework that integrates the process of \underline{\textbf{I}}nduction, \underline{\textbf{DE}}duction, and \underline{\textbf{A}}bduction. The \method agent generates initial hypotheses from limited observations through abduction, devises plans to validate these hypotheses or leverages them to solve problems via deduction, and refines previous hypotheses through induction, dynamically establishing and applying rules that mimic human rule-learning behaviors. Our evaluation of the \method framework, which involves five representative LLMs, demonstrates significant improvements over the baseline. 
Furthermore, our study with human participants reveals notable discrepancies in rule-learning behaviors between humans and LLMs.
We believe our benchmark will serve as a valuable and challenging resource, and \method will provide crucial insights for the development of LLM agents capable of human-like rule learning in real-world scenarios. Our code and data have been released on GitHub.\footnote{\href{https://github.com/KaiyuHe998/RULEARN_IDEA}{https://github.com/KaiyuHe998/RULEARN\_IDEA}}
\end{abstract}

\section{Introduction}
\begin{figure}[t]
    \includegraphics[width=0.9\columnwidth]{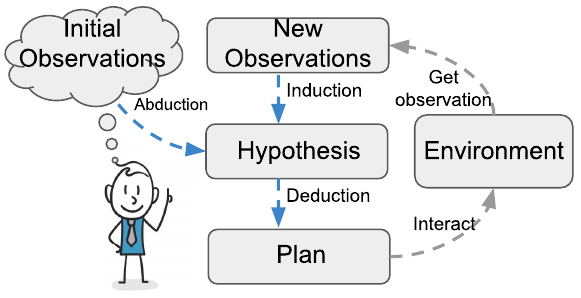}
    \vspace{-0.7em}
    \caption{The reasoning cycle of rule learning encompasses abduction, deduction, and induction.}
    \vspace{-2em}
    \label{fig:reasoning_introduction}
\end{figure}

One major pillar of human intelligence is the ability to discern and apply rules. We identify patterns, form hypotheses, and iteratively refine them through interaction with the environment—a process that traditionally involves three stages: abduction, deduction, and induction. According to Charles Peirce \citep{6714b41d-9d0c-30e9-8b06-a147d30ad866, peirce1974collected}, rule learning begins with an explanatory hypothesis \textbf{(abduction)}, followed by iterative experiments \textbf{(deduction)} and hypothesis refinement \textbf{(induction)} (see Figure~\ref{fig:reasoning_introduction}). This interdependent process underpins human rule learning in the real world, yet recent studies often isolate these stages in non-interactive settings \citep{bowen2024comprehensive, wang2023hypothesis, saparov2024testing, liu2024incomplete}.
\begin{figure*}
    \centering
    \includegraphics[width=1\textwidth]{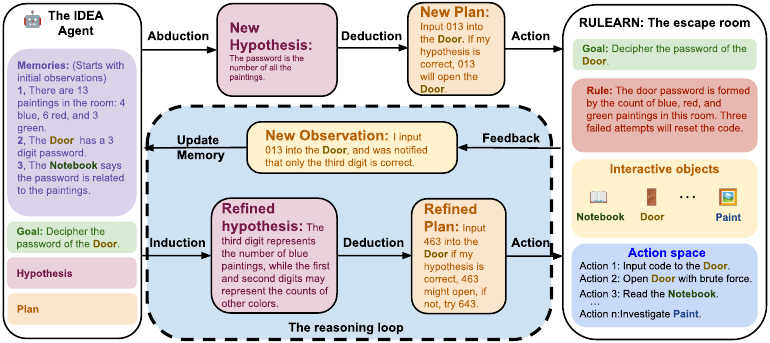} 
    \caption{\small A simplified puzzle in the \bench benchmark and the \method agent's workflow (in real puzzles, agents have fewer initial observations and more complex rules). The agent generates an initial hypothesis through abduction, develops an exploration plan via deduction, and refines its hypothesis using induction. For example, the \method agent first hypothesizes that the password is the number of the blue paintings, tests this by entering code 003, and adjusts its strategy based on the feedback. }
    \label{Fig:RULEARN_IDEA_introduction}
    \vspace{-1em}
\end{figure*}

To simulate the full complexity of human rule learning, three essential principles must be met: an \textbf{interactive environment} that encourages dynamic experimentation, a \textbf{fine-grained action space} that enables flexible and precise inputs for nuanced hypothesis testing, and the presence of \textbf{unknown rules} that force agents to infer, test, and revise hypotheses based solely on sparse observations. By integrating these three key principles, we introduce \bench, which features 300 high-quality, manually created puzzles with hidden rules set in a text-based environment, where agents begin exploration without any prior knowledge of the underlying rules. \bench simulates human-like rule learning—compelling agents to experiment dynamically, utilize fine-graind actions for detailed hypothesis testing, and infer rules from limited data.

Successfully solving the puzzles requires the agent to strategically select actions, efficiently gather pattern-revealing observations, and accurately reason from them to infer the hidden rules. \bench presents substantial challenges, as agents must rely on observations generated by their chosen actions to discern rules; without clear patterns emerging from their exploration, the agents are likely to fail.

\bench consists of three types of environments to evaluate the rule-learning ability in different scenarios: (1) \textbf{The Function Operator:} Determining the coefficients of mathematical functions defined by hidden expressions. Agents can assign various values to the input variables and observe the outputs, using this information to hypothesize the function's form. The challenge lies in efficiently selecting input values that reveal the underlying structure and accurately computing the coefficients based on limited observations. (2) \textbf{The Escape Room:} Deciphering the passcode to exit an escape room. A hidden rule determines how the objects in the room infer the passcode digits. Agents interact with these objects to gather clues and input passcodes into the door. Based on feedback, agents formulate hypotheses and infer the relationship between the objects and the passcode. (3) \textbf{The Reactor:} Synthesizing target strings using a reactor with a hidden string-combining rule. Agents need to experiment with different inputs and analyze outputs to deduce the reactor's transformation rule and achieve the desired outcome. 

To tackle the challenge in \bench, we introduce \method, a novel reasoning framework that integrates the process of \underline{\textbf{I}}nduction, \underline{\textbf{DE}}duction, and \underline{\textbf{A}}bduction. The \method agent employs these reasoning processes iteratively to explore the environments, learn rules, and achieve goals. In the \textbf{abduction} phase, the \method agent generates an initial hypothesis from limited observations. During the \textbf{deduction} phase, the \method agent creates and executes plans to attempt objectives or test its hypothesis. In the \textbf{induction} phase, the \method agent refines its hypothesis based on new observations, enhancing their accuracy and robustness. This iterative cycle enables the LLM agent to continually improve the learned rules through environmental feedback. An overview of how the \method agent solves puzzles in \bench is shown in Figure~\ref{Fig:RULEARN_IDEA_introduction}.

We evaluate \method on five popular LLMs—GPT-3.5-Turbo, GPT-4o, Gemma-7B, Llama3-8B, and Llama3-70B—observing roughly a 10\% improvement in success rates compared to the baseline. Without hypothesis guidance, the baseline agent tends to choose direct, uninformed actions that fail to uncover the hidden rules. In contrast, the \method agent reduces repeated actions by 30.2\%, obtains more diverse observations, and better understands the underlying rules. To further investigate their rule-learning capabilities, we compare LLM performance with that of 50 human participants. Although \method narrows the gap, LLMs still face challenges: (1) inefficient exploration in unfamiliar environments, resulting in insufficient evidence to reveal rules; (2) difficulty in deducing valid plans to verify current hypotheses and guide future exploration; and (3) reluctance to correct initial hypotheses when confronted with contradictory observations. These findings provide important insights into improving LLM agents to achieve more efficient rule learning in complex scenarios.

\section{Related Works}
\label{rule learning}

Agents powered by large language models (LLMs) have shown notable progress in understanding complex tasks \citep{Wang_2024,chen2023walkingmemorymazecontext, zhou2023recurrentgptinteractivegenerationarbitrarily, wang2024enhancinglargelanguagemodel,andreas2022languagemodelsagentmodels, park2023generativeagentsinteractivesimulacra, zhong2023memorybankenhancinglargelanguage,zhang2024surveymemorymechanismlarge,nakano2022webgptbrowserassistedquestionansweringhuman, lu2023chameleonplugandplaycompositionalreasoning, shi2023replugretrievalaugmentedblackboxlanguage,schick2023toolformerlanguagemodelsteach,yuan2023distillingscriptknowledgelarge, shen2023hugginggptsolvingaitasks,yao2023treethoughtsdeliberateproblem, Besta_2024}. Recent work examines different reasoning processes (abduction, deduction, induction) in LLMs \citep{bowen2024comprehensive, wang2023hypothesis, saparov2024testing, cheng2024inductivedeductiverethinkingfundamental, yang2024languagemodelsinductivereasoners}, but typically in isolation. As a result, their comprehensive rule-learning abilities in interactive settings remain underexplored.

\begin{table*}
\caption{\small The reacting rules in the Reactor Puzzle. All letters are functionally equivalent and exhibit no special behaviors. Identical symbols represent the same letter, while different symbols denote different letters. Each puzzle operates under one specific rule. The Middle Insertion rule inserts the shorter string into the longer string; if the length of the longer string is odd, the shorter string is inserted just to the right of the center. If both strings are of equal length, the second string is inserted into the middle of the first string. The Prefix Replacement rule retains the prefix of the longer string and concatenates it with the shorter string, dropping the tail of the longer string results in two output strings. There are two special cases where the strings are simply concatenated in order.}
\label{React rule table}
\begin{center}
\vspace{-1.2em}
\resizebox{1\textwidth}{!}{
    \small
    \begin{tabular}{lllcc} 
    \multicolumn{1}{c}{\bf Rule Description} &\multicolumn{1}{c}{\bf Example Reaction 1} &\multicolumn{1}{c}{\bf Example Reaction 2}&\multicolumn{1}{c}{\bf Special Case 1}&\multicolumn{1}{c}{\bf Special Case 2} \\ \hline \\
    Simple Concatenation  & AB + C = ABC & AB + CDE = ABCDE  & \textemdash & \textemdash\\
    Reverse Concatenation  & AB + C = CAB & AB + CDE = CDEAB & \textemdash & \textemdash\\
    Middle Insertion  & AB + C = ACB & AB + CDE = CDABE & A + B = AB & \textemdash\\
    Prefix Replacement & AB + C = AC + B & AB + CDE = CAB + DE & AB + CD = ABCD & AA + A = AAA
    \end{tabular}
}
\end{center}
\vspace{-1em}
\end{table*}

Current reasoning tasks are hindered by inadequate benchmarks that either rely on QA datasets like Hotpot-QA \citep{yang2018hotpotqadatasetdiverseexplainable} and Trivia-QA \citep{joshi2017triviaqalargescaledistantly}—which lack active information gathering—or by coarse-grained interactive environments such as TextWorld \citep{côté2019textworldlearningenvironmenttextbased} and AlfWorld \citep{shridhar2021alfworldaligningtextembodied} that limit agents to high-level actions (e.g., go to, open), impeding complex, experiment-driven rule discovery. This is in stark contrast to real-world rule learning, which requires active evidence gathering, experimentation, and iterative refinement. Moreover, many studies use static, non-interactive scenarios where LLMs receive all information upfront \citep{yang2023failurespavewayenhancing,zhu2024largelanguagemodelslearn,shi2023languagemodelsimproveevent,liu2024incomplete}, failing to capture the dynamic nature of real-world learning. Even recent efforts \citep{xu2024active, Montes_2022} that integrate interactivity treat information gathering, rule generation, and application as distinct phases, undermining the development of agents capable of seamlessly integrating these elements.

We claim that to fully capture rule‐learning ability in the real world, three criteria should be met: \textbf{(i) Interactive Environment:} the environment must be interactive so agents learn from interaction rather than passively receiving data; \textbf{(ii) Fine-graind Action Space:} the action space must be fine-grained, since existing benchmarks offer only coarse actions (e.g., go to, open) that prevent agents from performing the detailed experiments needed to test hypotheses \citep{jansen2024discoveryworldvirtualenvironmentdeveloping}; and \textbf{(iii) Unknown-Rule:} the target rules must be unknown, as in ScienceWorld \citep{wang2022scienceworldagentsmarter5th} most test cases rely on knowledge LLMs have already mastered during pretraining. To address this gap, our proposed \bench fulfills all three requirements by providing a fully interactive environment, offering a fine-grained action space in which agents submit arbitrary strings that our system parses to deliver character-level feedback, and introducing new, manually crafted puzzles whose rules are not familiar to LLMs. In turn, \method equips agents to manage the interdependent processes of information gathering, hypothesis generation, and validation within a unified framework that closely mimics human rule-learning behavior.

\vspace{-0.7em}
\section{The \bench Benchmark}
\vspace{-0.7em}

\label{CHIBI-100 Benchmark}
We develop three puzzle sets—\textbf{Function Operator}, \textbf{Escape Room}, and \textbf{Reactor}—each consisting of 100 unique, manually created puzzles of varying complexity, with each set reflects a different real-world rule-learning scenario.
Unlike existing fine-grained interactive environments, which are predominantly found in the robotics domain and offer significantly fewer tasks \citep{jain2020cordialsyncgoingmarginal, nasiriany2024robocasalargescalesimulationeveryday, zhang2024buildingcooperativeembodiedagents}, \bench is the first text-based environment providing such fine-grained interactions specifically for language agents.

\textbf{The Function Operator.} This puzzle type simulates scenarios where systemic theories or established knowledge (e.g., mathematics) are applicable for efficiently testing hypotheses. The agent interacts with a set of univariate multi-term equations involving integer parameters from [0,9] and elementary functions of the variable $x$, selected from $f(x) \in \{x^0, x^1, x^2, \sin(x), \frac{1}{x}, |x|, -x\}$. The agent is provided with the number of functions, the presence of specific parameters in each function (the exact numerical values of these parameters are unknown and represented by letters), and the types of elementary functions involved in the current puzzle. The goal of the agent is to deduce the values of these parameters. For example, in one puzzle, the ground truth is $\mathbf{F_1}(x) = a \sin(x) + b \times \frac{1}{x}$, $\mathbf{F_2}(x) = a x^2$ where $a=3$ and $b=2$. The agent knows the following information: There are three elementary functions in this puzzle $\{\sin(x), \frac{1}{x}, x^2\}$, there are two functions $\mathbf{F_1}(x)$ and $\mathbf{F_2}(x)$, $\mathbf{F_1}(x)$ has 2 terms and parameters $a, b$ in it, and $\mathbf{F_2}(x)$ has 1 term and one parameter $a$.  To solve the puzzle, the agent must interact with the environment through a defined action space: selecting a function and assigning values to $x$, then observing the resultant output. For example, assigning values 1 and 2 to $\mathbf{F_2}$ reveals a quadratic increase in output, indicating the presence of $x^2$ in $\mathbf{F_2}$. Similarly, assigning a value of 1 to $\mathbf{F_1}$ results in a floating-point output, rather than an integer, suggesting the inclusion of trigonometric components, confirming that $\sin(x)$ is a component of $\mathbf{F_1}$. The difficulty of each puzzle is controlled by variations in the number of functions, unknown parameters, and elementary functions in use. We manually enumerated 100 combinations of functions, incorporating different numbers of terms and types of elementary functions to ensure a diverse range of puzzle complexity (see detailed distribution in Table~\ref{tab:Function operator distribution} in Appendix ~\ref{Appendix tables}). 

\textbf{The Escape Room.} This environment simulates scenarios where no established knowledge is applicable, challenging agents to rely on basic human priors—such as counting, mapping, and attribute abstraction—to convert qualitative observations into general rules through iterative feedback. We create a fictitious setting: an agent is placed in an art gallery escape room and must decipher a 3-digit password to unlock a code-secured door. Each digit of the password represents the count of paintings of a specific \textbf{type}—watercolor, oil, or acrylic—that share a given \textbf{color}. The agent receives brief descriptions of paintings, such as \textit{``This is an acrylic painting of a green jungle''}, indicating their type and color. Initially, the agent only knows the password is a 3-digit number and is given a hint about which color to focus on. After proposing a hypothesis and entering a password guess, the door provides feedback on which digits are correct, allowing the agent to refine its hypothesis. To prevent a brute-force approach, the specific color associated with the password changes after every three failed attempts. Each puzzle varies in the number of paintings, and while paintings in the same room are visible, those in other rooms remain hidden until the agent moves to access them. The difficulty of this puzzle type is controlled by the different number of paintings, whether agent need to  as detailed in Table~\ref{tab:Escape room distribution} in the Appendix ~\ref{Appendix tables}.

\textbf{The Reactor.} This environment simulates scenarios without pre-established knowledge, requiring agents to perform sequential, interdependent actions to uncover ordering and transformation patterns—mirroring real-world experimental design, where each step influences the next. Specifically, the agent's task is to synthesize target strings using a reactor governed by a hidden string-combining rule. These strings are represented by sequences of alphabetic letters, such as \textit{A}, \textit{B}, \textit{AABB}, and \textit{CAB}. The reactor permits the agent to input two strings, initiating a reaction that produces a new string for use in subsequent experiments. The agent's objective is to decipher the specific rules that govern string synthesis by methodically testing different string combinations, with the ultimate goal of synthesizing the target string using the discovered rules. We have designed four types of rules, detailed in Table~\ref{React rule table}. The difficulty of this puzzle type is controlled by the specific rules used, the length of the target string to be synthesized and the number of unique letters contained in the target string, as detailed in Table~\ref{tab:Reactor distribution} in Appendix~\ref{Appendix tables}.

Together, these puzzle types simulate a broad spectrum of real-world rule learning by requiring agents to apply both formal knowledge and commonsense reasoning. Detailed statistics for each puzzle type and example puzzles are provided in Appendix~\ref{Appendix tables} and~\ref{Puzzle Examples}. The \bench benchmark is designed to emulate realistic, complex text environments with diverse rules. To preserve this realism, we do not restrict rule representations to a specific formal language; instead, LLM agents use natural language to describe rules, promoting generalizability and preventing prior knowledge that could undermine the challenge.


\section{The \method Agent}
\label{The IDEA agent}

\begin{figure*}[ht]
    \centering
    \includegraphics[width=1\textwidth]{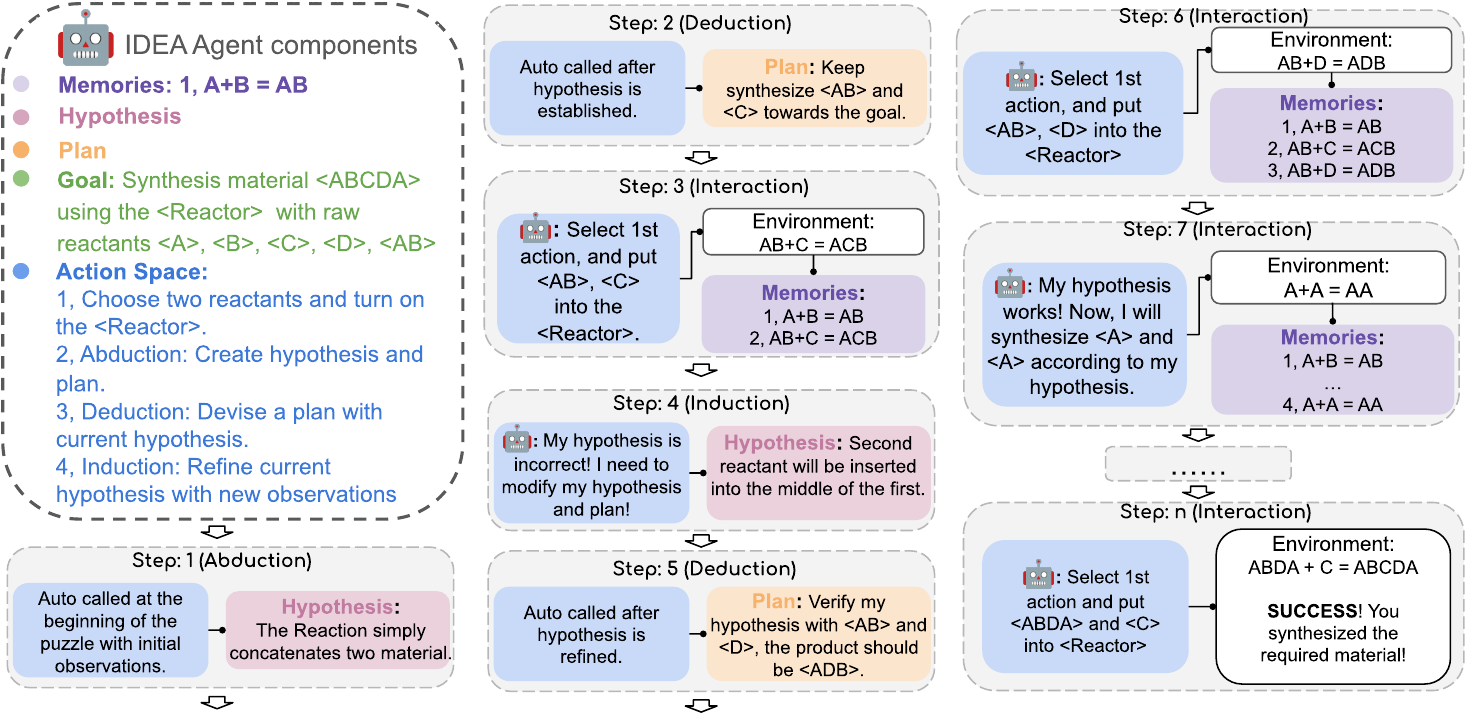}
    \vspace{-2em}
    \caption{\small An example of the \method agent solving a Reactor puzzle. At each step, the agent must choose whether to interact with the environment or adjust its hypothesis and plan based on current observations. If observed facts contradict the existing hypothesis, the agent is expected to refine its hypothesis. The refined hypothesis and plan will then guide subsequent exploration.}
    \label{fig:Pipeline_plot}
    \vspace{-1em}
\end{figure*}

We introduce \method, a novel reasoning framework that integrates the process of \underline{\textbf{I}}nduction, \underline{\textbf{DE}}duction, and \underline{\textbf{A}}bduction to learn rules in interactive environments.
The \method agent consists of the following components: Goal(G), Action Space($\mathbb{A}$), Memory($\mathbb{M}$), Hypothesis(H), and Plan(P), which are elaborated in Table~\ref{Method:components}. 

Upon beginning to explore a puzzle, we initialize the agent memory with an initial observation of the environment. The agent's goal is initialized with the objective of the puzzle, e.g., \textit{synthesize a target string} for a Reactor puzzle. The agent's action space is initialized as the set of interactive actions defined by the puzzle, such as \textit{choosing two strings and running the Reactor}, as well as establish the initial hypothesis (abductive action), devises a plan to validate or leverage hypothesis (deductive action), and refining the current hypothesis (inductive action).  

The \method agent begins with an abductive action to generate an initial hypothesis, followed by a deduction step to create a new plan. Based on this plan, the agent interacts with the environment. Upon receiving feedback from the environment as a new observation, the agent may take an inductive action to refine the hypothesis or perform another interaction with the environment. Deductive action is invoked to adjust the plan every time the hypothesis changes. This reasoning loop continues until the puzzle is solved or a maximum number of steps is reached. After each step, the results are appended to the agent's memory, including interaction outcomes and any modifications to the hypothesis or plan. We provide a simplified algorithm demonstrating how the \method agent operates in Algorithm~\ref{simplified_algo}. Specifically, at each step, we prompt the LLM to reflect on the information recorded in the \method agent's components to make decisions and take actions. Detailed prompts for each type of action are available in Appendix~\ref{Prompt example}.


More detailed implementation of the agent can be found in Appendix~\ref{appendix:Environment Entities}. 
Similar to real-life scenarios, when agents solve tasks in \bench puzzles, they do not know the outcomes in advance. Consequently, it is challenging to decide when to refine or change their hypothesis and plans, as well as what interactive actions to take to gather pattern-revealing observations. A detailed example of the \method agent solving the Reactor Puzzle is provided in Figure~\ref{fig:Pipeline_plot}.

\begin{table}
\vspace{-1.9em}
\centering
\begin{algorithm}[H]
\caption{\method Agent Rule-learning Loop}
\label{simplified_algo}
\begin{algorithmic}[1]
\footnotesize
\Procedure{RuleLearningLoop}{}
    \State Initialize Goal(G), Action Space$\mathbb{(A)}$
    \State $\text{Memory}\mathbb{(M)} \gets \text{Initial observations}$
    \State $\text{\#step} \gets 0$
    \State $\text{Hypothesis(H)} \gets \textbf{Abduct}(\text{G},\mathbb{A,M})$
    \State $\text{Plan(P)} \gets \textbf{Deduct}(\text{H},\text{G},\mathbb{M,A})$
    \State $\mathbb{M}.\text{add}(\text{``New hypothesis and plan''}, \text{H}, \text{P})$
    \While{$\text{\textbf{G} not achieved and } \text{\#step} \leq \text{max\_step}$}
        \State $\mathbf{a} \gets \text{select\_action(\text{G, H, P, $\mathbb{M,A}$})}$ 
        \If{$\mathbf{a}$ is interactive action}
            \State $\text{result} \gets \text{execute\_action}(\mathbf{a},\text{G},\text{H},\text{P},\mathbb{M})$
            \State $\mathbb{M}.\text{add(result)}$
            \State $\text{\#step} \gets \text{\#step} + 1$
        \ElsIf{$\mathbf{a}$ is inductive action}
            \State $\text{H} \gets \textbf{Induct}(\mathbf{a},\text{G},\mathbb{M},\text{H},\text{P})$
            \State $\text{P} \gets \textbf{Deduct}(\text{H},\text{G},\mathbb{M,A})$
            \State $\mathbb{M}.\text{add}(\text{``Refined hypothesis and plan''}, \text{H}, \text{P})$
        \EndIf        
    \EndWhile
\EndProcedure
\end{algorithmic}
\end{algorithm}
\end{table}

\begin{table}
\label{idea_componenets}
\begin{center}
\footnotesize
\vspace{-1.5em}
\resizebox{0.48\textwidth}{!}{
    \begin{tabular}{lp{5cm}} 
    \toprule
    \textbf{The \method Agent Component} & \textbf{\makecell[l]{Definition}}\\
    \midrule
    Goal(G) & Goal of the agent in the current puzzle.\\
    \midrule
    Action Space($\mathbb{A}$) & Set of actions the agent can take, including abductive action, deductive action, inductive action, as well as the set of interactive actions defined by the puzzle. \\
    \midrule
    Memory($\mathbb{M}$) & Set of natural language strings to record all interaction results till the current step. \\
    \midrule
    Plan(P) & Generated plans to guide future actions.\\
    \bottomrule
    \end{tabular}
}
\vspace{-1.2em}
\end{center}
\caption{\small Components of the \method agent.}
\label{Method:components}
\vspace{-1.2em}
\end{table}

\section{Experiment Results}
\subsection{Experiment Settings}

To evaluate the effectiveness of \method, we respectively initialize it with three popular open-source LLMs, including Gemma-7B \citep{geminiteam2024geminifamilyhighlycapable}, Llama3-8B, and Llama3-70B \citep{dubey2024llama3herdmodels}, and two closed-source LLMs, GPT-3.5-Turbo \citep{dubey2024llama3herdmodels} and GPT-4o\footnote{\href{https://openai.com/index/gpt-4o-system-card/}{https://openai.com/index/gpt-4o-system-card/}}. We compare \method against the following two variants: 

\begin{itemize}[itemsep=-1pt,topsep=3pt,leftmargin=10pt]
    \item \textbf{ReAct Agent (Baseline)}: We choose ReAct \citep{yao2023reactsynergizingreasoningacting} as our baseline. The ReAct agent does not incorporate the full reasoning loop of abduction, deduction, and induction, nor does it generate explicit hypotheses or plans. Instead, at each step, it reasons over its current memories and the goal and selects an interactive action accordingly.

    \item \textbf{Oracle-rule Agent}: Even if the agent could successfully learn the correct rule, applying the learned rule to solve the puzzle is non-trivial. The Oracle-rule agent serves as a control group to establish the Oracle performance with the ground-truth rule provided. Specifically: 1) For the Function Operator puzzles, agents are given the exact forms of the functions. Their task is to derive the values of the coefficients. 2) For the Escape Room puzzles, agents are provided with how the password is constructed from the objects. Their task is to derive the password using the provided rule. 3) For the Reactor puzzles, the reaction rule is given to the agents in natural language accompanied by examples. The agents only need to synthesize the target strings.

\end{itemize}

Each variant is evaluated across all three puzzle types. We set the temperature for LLMs to 0, based on observations that models like GPT-4o perform better at lower temperatures. The prompts used for the agents are detailed in Appendix~\ref{Prompt example}. Additionally, since the success rate does not improve after 15 interactive steps for LLMs, we capped the maximum interaction step count at 15. An agent is considered to have failed a puzzle if it does not solve it within these 15 steps. Details on the computational budget are available in Appendix~\ref{comp_budget}.

\subsection{Human participants}

To compare human and LLM performance in abduction, deduction, and induction reasoning, we recruited 50 participants and assigned each three randomly selected puzzles (10 from each reasoning type). Each puzzle was attempted by five different participants, with no prior exposure to the rules. Participants followed the same reasoning procedure outlined in \method, which mirrors their natural problem-solving methods and does not bias their responses. They documented their reasoning processes, enhancing transparency and facilitating clearer comparisons with LLMs. Further details on IRB approval and participant recruitment are in \S\ref{ethics statement}. Attempts failing to solve a puzzle within 15 steps were marked as unsuccessful, ensuring fair comparisons. A sample user interface is shown in Figure~\ref{fig:user interface} in Appendix~\ref{user interface section}.

\subsection{Main Results}

We calculated average puzzle solving success rate across different variants. The detailed results are displayed in Table~\ref{tab:success_rate_table}.
\begin{table*}
\caption{\small Puzzle Success Rate. The success rates for each setting. Across all LLMs, \method achieves consistently significant improvements, except for Gemma-7B in the Reactor puzzles. We use boldface to highlight performance comparisons between the Baseline and IDEA agents with GPT-4o.}
\vspace{-1em}
\label{tab:success_rate_table}
\begin{center}
\small
\resizebox{\textwidth}{!}{
\begin{tabular}{llcccc} 
\hline
\textbf{Setup} & \textbf{LLMs} & \textbf{All Types (\%)} & \textbf{Function Operator (\%)} & \textbf{Escape Room (\%)} & \textbf{Reactor (\%)} \\ 
\hline
\multirow{5}{*}{\textbf{Oracle-rule Agent}} 
& Gemma-7B  & $1.67$ & $0.0$ & $5.0$ & $0.0$\\
& Llama3-8B  & $5.67$ & $1.0$ & $14.0$ & $2.0$\\
& Llama3-70B & $32.67$ & $33.0$ & $48.0$ & $17.0$\\
& GPT-3.5-Turbo & $6.33$ & $7.0$ & $11.0$ & $1.0$\\
& GPT-4o & $66.0$ & $77.0$ & $91.0$ & $30.0$\\
\hline
\multirow{5}{*}{\textbf{ReAct Agent (Baseline)}}
& Gemma-7B  &$0.33$ & $0.0$ & $0.0$ & $1.0$\\
& Llama3-8B  & $1.67$ & $0.0$ & $5.0$ & $0.0$\\
& Llama3-70B  & $19.67$ & $33.0$ & $17.0$ & $9.0$\\
& GPT-3.5-Turbo & $5.33$ & $13.0$ & $3.0$ & $0.0$\\
& GPT-4o & $\mathbf{43.33}$ & $\mathbf{62.0}$ & $\mathbf{45.0}$ & $\mathbf{23.0}$\\
\hline

\multirow{5}{*}{\textbf{\method Agent (Ours)}}
& Gemma-7B  & $0.33$ & $0.0$ & $1.0$ & $0.0$\\
& Llama3-8B  & $4.33$ & $7.0$ & $5.0$ & $1.0$\\
& Llama3-70B & $29.0$ & $41.0$ & $35.0$ & $11.0$\\
& GPT-3.5-Turbo & $7.33$ & $18.0$ & $3.0$ & $1.0$\\
& GPT-4o & $\mathbf{50.33}$ & $\mathbf{73.0}$ & $\mathbf{51.0}$ & $\mathbf{27.0}$\\
\hline
& Human  & $63.33$ & $66.0$ & $56.0$ & $68.0$\\
\hline
\end{tabular}}
\vspace{-1.7em}
\end{center}
\end{table*}

For the Oracle-rule agent, 
in the Escape Room puzzles, agents achieve up to an 89\% success rate by simply following the provided rule. However, in other puzzles, merely knowing the rule is not sufficient for success; applying the rule to solve the puzzle remains challenging. 
The Baseline agent is not provided with the underlying rules and solely relies on historical observations to make interactive actions. 
Across models, the success rates drop by about half compared to the Oracle-rule agent. This significant decrease highlights the challenge of rule learning and indicates that current LLM agents struggle to learn rules in unfamiliar environments without explicit guidance.

\noindent \textbf{\method significantly boosts success rates.} Our proposed \method framework leads to approximate 10\% increases in success rates for Llama3-70B, GPT-3.5-Turbo, and GPT-4o compared to the Baseline agent. This improvement demonstrates that incorporating a reasoning loop of abduction, deduction, and induction substantially enhances the LLM rule-learning performance in unfamiliar environments. 
\method enables the LLMs to generate hypotheses, plan actions, and refine their understanding based on new observations, which is crucial for rule learning. 
However, smaller models like Llama3-8B and Gemma-7B do not perform better when applying \method. Small models inherently struggle with complex tasks like \bench—even when given the ground truth rule, their performance remains near 0—so no agent framework can significantly boost their performance.

\noindent \textbf{LLM agents still fall far behind humans.} In the Escape Room puzzle, where the primary challenge is to discover the rule, the Oracle-rule agents excel because once the rule is identified, applying the rule is simple. However, in other types of puzzles, human participants significantly outperform all LLM agents, including the Oracle-rule agents, even without knowing the rules beforehand. 

\subsection{Analysis}
\begin{figure}[t]
  \centering
  \includegraphics[width=\columnwidth]{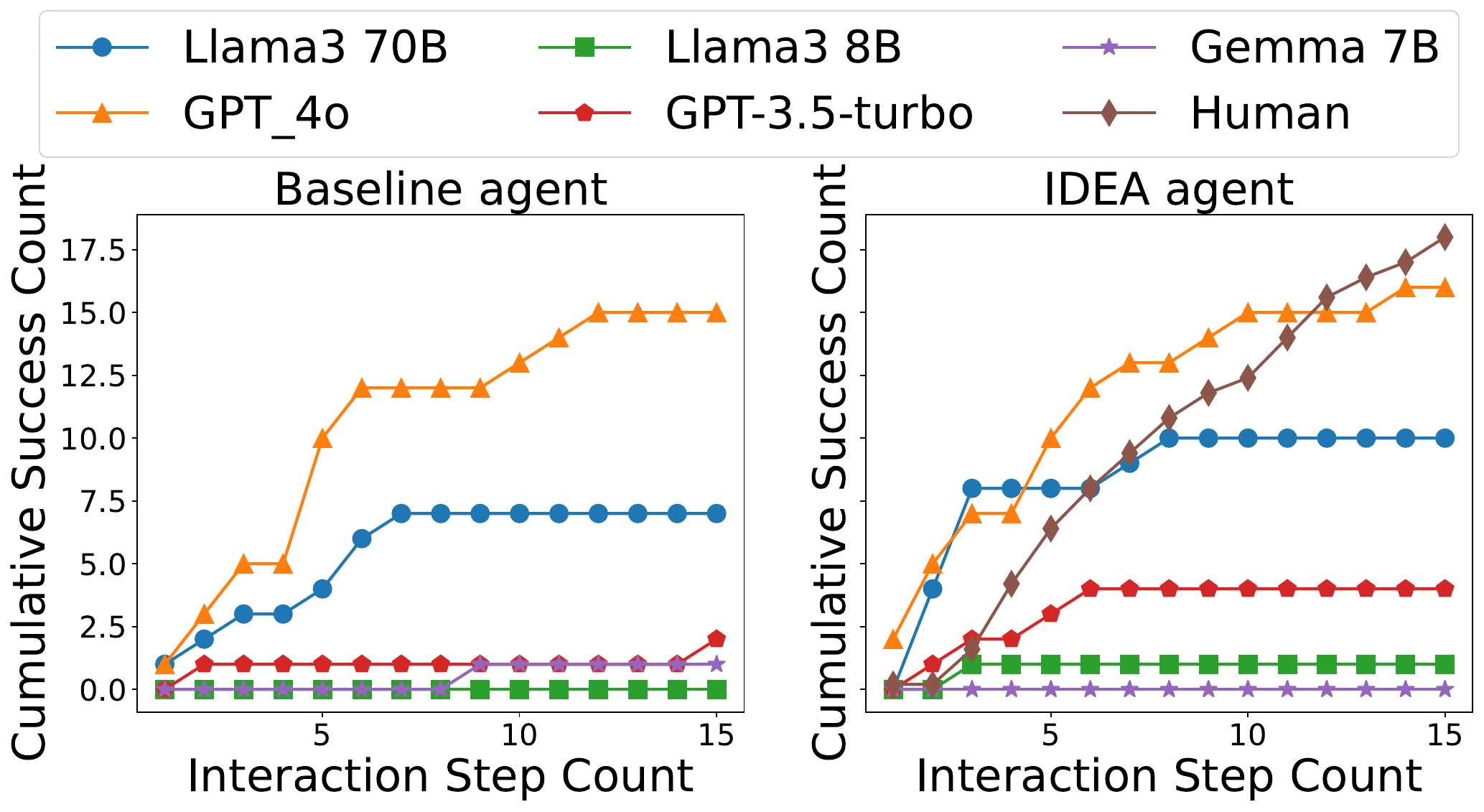}
  \caption{\small Comparison of the cumulative number of puzzles solved at each interaction step. The \method agent significantly decrease number of steps needed to solve a puzzle compared to the Baseline agent.}
  \vspace{-1.5em}
  \label{fig:solving_speed}
\end{figure}

\noindent \textbf{\method solves puzzles with fewer steps.} Figure~\ref{fig:solving_speed} illustrates the cumulative number of puzzles solved at each interaction step for the Baseline agent, the \method agent, and human participants. The slopes of the lines represent the rate at which puzzles are solved per step. Compared to the Baseline agent, the \method agent exhibits a steeper slope, indicating that the integration of abductive, deductive, and inductive reasoning enhances the agent's efficiency in exploring the environment and learning the underlying rules at each interactive step, especially during the early stages.

When focusing on human participants, we observe that they solve fewer puzzles in the initial steps. However, as interactions continue, the number of puzzles solved by humans increases rapidly, eventually surpassing that of all LLM agents. In contrast, LLM agents solved 88.76\% of the puzzles within the first 10 steps. Beyond this point, additional interactions contribute less to their success rate. This pattern suggests that humans have a superior ability to learn continuously from interactive environments, effectively improving their performance over time. If we did not limit the puzzles to 15 steps, we anticipate that the success rate of human participants would be even higher. 
\begin{table*}[t]  
\caption{\small Ablation on task-agnostic vs.\ task-specific prompts (GPT-4o, 50 \% puzzle subset).}
\vspace{-1em}
\label{tab:agnostic_ablation}
\centering
\small
\resizebox{\textwidth}{!}{   
\begin{tabular}{lcccc}
\hline
\textbf{Setup} & \textbf{All Types (\%)} & \textbf{Function Operator (\%)} & \textbf{Escape Room (\%)} & \textbf{Reactor (\%)} \\
\hline
Oracle-rule Agent          & 67.23 & 75.16 & 91.94 & 29.03 \\
ReAct Agent (Baseline)     & 34.30 & 50.00 & 33.22 & 24.83 \\
IDEA Agent (Task-Specific)      & \textbf{45.30} & \textbf{62.88} & \textbf{45.80} & \textbf{24.84} \\
IDEA Agent (Task-Agnostic) & \textbf{50.00} & \textbf{75.16} & \textbf{45.80} & \textbf{29.03} \\
\hline
\end{tabular}}
\vspace{-1.7em}
\end{table*}

\noindent \textbf{\method reduces repetitive actions.} LLM agents frequently repeat previous actions instead of exploring new ones. This behavior is highly inefficient in our controlled puzzle environments, where each interaction yields deterministic results, and repeating the same action generally does not provide new information. 
We calculate the average number of repeated actions performed while solving each puzzle, with detailed statistics in Table~\ref{tab:repeat_count} in Appendix~\ref{Appendix tables}. We observe that most LLMs commonly repeat actions in the Baseline agent. 
The \method agent effectively reduces this tendency by explicitly generating plans during the deduction phase. By outlining a clear plan, the \method agent can better assess whether the current observations are sufficient or if further specific evidence is needed to reveal the underlying rule. For example, in the Escape Room puzzle, the \method agent avoids unnecessary attempts at entering passwords when the evidence gathered is sufficient to determine the correct code (see Figure~\ref{fig:repeatition example} in Appendix~\ref{Prompt example}). 

\noindent \textbf{\method relies on the reasoning ability of underlying LLMs.}
The effectiveness of \method depends on the underlying LLMs' ability to reason from hypotheses and observations. Particularly, if an agent generates a false hypothesis and fails to properly refine it, being guided by this incorrect hypothesis can lead the agent to perform even worse than the baseline. During our experiments, we observed that current LLMs tend to hallucinate, especially in the Escape Room puzzles and more severely in the Reactor puzzles. This results in smaller performance improvements compared to those seen with the Function Operator puzzles. This is likely because such fictitious scenarios are not extensively represented in LLM training data. Moreover, LLMs struggle to recognize letter-level patterns, and their reasoning capabilities still require significant enhancement. Examples of hallucination can be seen in Appendix~\ref{Hullicination examples}). 

\noindent \textbf{\method is robust to different prompts.} Recent work shows that large deep models often rely on shallow pattern matching and thus are sensitive to prompt changes and unable to generalize to new environments \citep{kang2024farvideogenerationworld, niu2024largelanguagemodelscognitive, sun2024instructionfollowingevaluatinginferential, mirzadeh2024gsmsymbolicunderstandinglimitationsmathematical}. For example, \citet{mirzadeh2024gsmsymbolicunderstandinglimitationsmathematical} report that changing an irrelevant word in a math problem, for instance replacing ``erasers'' with ``notebooks,'' can greatly affect an LLM’s prediction.

As \method is expected to operate and learn new knowledge by interacting with unfamiliar environments, we conducted a supplementary experiment to test whether it remains effective under task-agnostic instructions. In our main experiments, we use task-specific instruction such as \textit{``Hypothesize the actual forms of each function.''} In the supplementary experiment, we replaced that phrase with \textit{``Please consider the given observations and propose an initial hypothesis that explains them.''} This prompt can apply equally to all rule-learning tasks (see Appendix~\ref{Task-Agnostic Prompts}). We randomly sampled half of our puzzles and re-ran the evaluation with this new instruction. As shown in Table~\ref{tab:agnostic_ablation}, \method still outperforms the baseline by a similar margin. Compared with other popular prompting method such as chain-of-thought (CoT) prompting \citep{NEURIPS2022_9d560961}, which typically requires carefully tailored prompts for each problem type, our single, high-level prompt works across all rule-learning tasks without further tuning. The agent first reviews its observations and then executes an abduction–deduction–induction loop, consistently delivering a performance boost.

\vspace{-0.5em}

\section{Fine-grained Human Evaluation}
\vspace{-0.5em}
\label{Section human evaluation}

\begin{figure*}[!ht]
    \centering
    \includegraphics[width=0.96\textwidth]{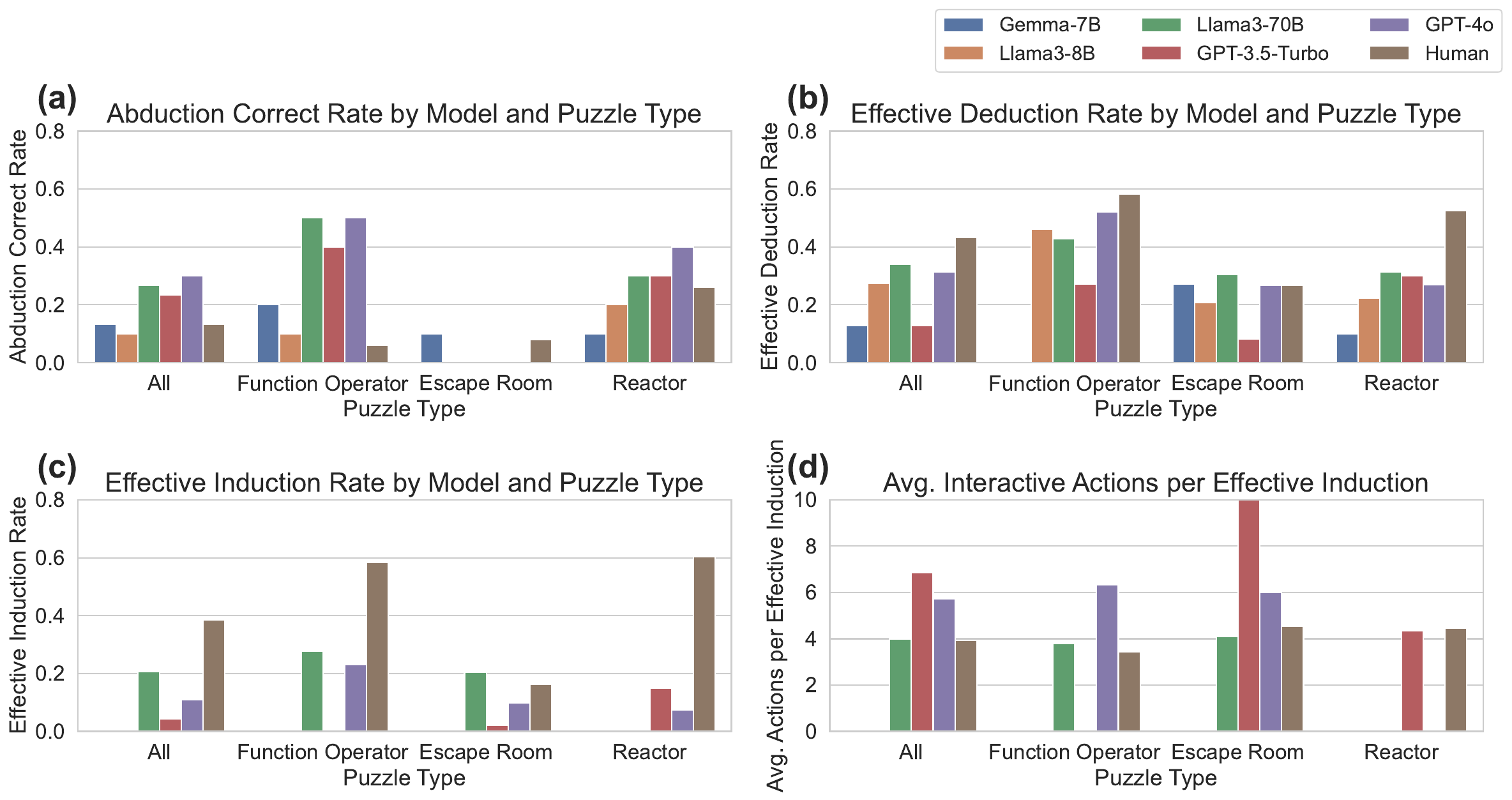}  
    \vspace{-0.8em}
    \caption{\small Human Evaluation Results. Bars represent measured values per model and puzzle type; the absence of a bar indicates zero or unavailable data. Plot (a): Abduction Correct Rate—the frequency of correctly guessing the rule during abduction. Plot (b): Effective Deduction Rate—the rate at which deduction plans effectively validate hypotheses or solve puzzles. Plot (c): Effective Induction Rate—the proportion of inductions where the refined hypothesis improved over the previous one. Plot (d): Average Actions per Effective Induction—the average number of interactive actions needed for an effective induction.}
    \label{fig:human_evaluation_result}
\end{figure*}
To compare LLM agent rule-learning with human processes, we conducted a fine-grained evaluation of rule-learning trajectories at every reasoning step. Computer science graduate students assessed the hypotheses and plans generated by both the \method agent and human participants during the abduction, deduction, and induction stages on a randomly selected 50\% subsample of puzzles.

\noindent \textbf{Abduction stage.} In this stage, agents formulate an initial hypothesis based on early observations—sometimes even guessing the ground truth rule in simpler puzzles. Figure~\ref{fig:human_evaluation_result}(a) indicates that LLMs such as GPT-4o correctly identify the rule during abduction about 30\% of the time, while only around 10\% of human participants generate a correct hypothesis under initial uncertainty. This discrepancy suggests that deviations in rule learning between LLMs and humans emerge as early as the abduction stage. LLMs tend to process every word of the prompt and produce a hypothesis even when unsure, whereas humans generally refrain from formalizing hypotheses under uncertainty.

\noindent \textbf{Deduction stage.} After establishing an initial hypothesis—or each time the agent refines its hypothesis—the agent derives a plan to validate it or attempt the puzzle. As shown in Figure~\ref{fig:human_evaluation_result}(b), humans generally outperform LLMs in creating high-quality plans. These superior plans enable humans to take a wider variety of actions, gathering more useful observations. According to Table~\ref{tab:repeat_count} in Appendix~\ref{Appendix tables}, humans ultimately collect 20\% more diverse observations compared with LLMs.

\noindent \textbf{Induction stage.} Figure~\ref{fig:human_evaluation_result}(c) shows the effective induction rate—the proportion of refined hypotheses that improve upon the previous version. Induction is crucial for developing high-quality hypotheses, and humans excel at this stage, with 40\% of their refined hypotheses showing improvement. In contrast, LLMs struggle to converge on the correct rule, with effective induction rates below 20\%. Moreover, they often fail to recognize conflicts between observations and hypotheses—for example, Llama3-70B rarely engages in induction within Reactor puzzles(see Appendix~\ref{Hullicination examples}).—resulting in redundant observations and fails the puzzle.

\noindent \textbf{Average interactions needed for effective induction.} Figure~\ref{fig:human_evaluation_result}(d) shows that humans require fewer interactions—approximately four on average—to effectively refine their hypotheses, compared to LLMs. While LLMs can process initial information thoroughly and generate plausible hypotheses, they face challenges in refining these hypotheses based on new observations during interaction with the environment (see Figure~\ref{fig:rule_found_rate_figure} in Appendix~\ref{Appendix Figures}). This limitation suggests that LLMs may struggle to learn from new observations and incorporate feedback to continuously improve their hypotheses and problem-solving strategies. This gap may become more pronounced when agents are faced with larger action spaces and more complex rules.

\vspace{-0.7em}
\section{Conclusion}
\vspace{-0.6em}
In this work, we introduce \bench, the first benchmark that (i) places LLM agents in fully interactive environments, (ii) gives them a fine-grained action space for discovering and applying rules, and (iii) employs manually designed rules that are unseen during pre-training. We propose \method, an agent framework that mimics human reasoning through abduction, deduction, and induction. Comprehensive experiments involving five prominent LLMs and human participants reveal that while \method significantly improves the rule learning ability of LLM agents, there is still a large gap between LLM and humans particularly in refining hypotheses and adapting strategies. Despite these advancements from the \method framework, LLMs still face challenges in generating valid hypotheses and avoiding repetitive actions in complex scenarios. Our findings underscore the need for further development of LLMs that can emulate human cognitive processes more effectively in explorations of novel environments. \bench provides a foundational resource for future research aimed at closing these gaps.
 
\section{Limitations}
While solving puzzles, the \method agent needs to manage long contexts. As exploration progresses and the agent encounters more observations, it must simultaneously process all observations. This requirement can limit its effectiveness in scenarios that involve lengthy contexts and complex rules, where extensive experimentation is needed to uncover these rules. By prioritizing and focusing on more critical observations, we can enhance the \method agent's performance in managing long-context scenarios and in tackling challenging puzzles that require multiple steps to gather sufficient evidence.

\section{Ethics Statement}
\label{ethics statement}
Our work aims to benefit the broader research community by introducing \bench, a benchmark for evaluating the rule-learning abilities of LLM agents and proposing the \method agent framework. All data in \bench contains no personal or sensitive information, ensuring respect for privacy and ethical standards. This project is approved by our Institutional Review Board (IRB). Human participants are recruited through emails from our university's computer science and engineering department. All participants were adults over 18 years old and provided informed consent. The data collected from these participants were de-identified and consented for release for research purposes. Participants were compensated \$15 each for one hour of their time. We ensured that all content presented during evaluations was free from offensive or inappropriate material. For human evaluations of all the hypotheses and plans generated by LLM agents and human participants, three computer science graduate students (our co-authors) conducted the evaluation. We are committed to the ethical use of our benchmark and agent framework, and upon acceptance of this paper, we will release our code and data to encourage open collaboration and advancement in the field.

\section{Acknowledgements}
We thank the anonymous reviewers for their thoughtful comments. We are also grateful to Jia Li, Kevin Wang, and Kangshuo Li for their valuable feedback on puzzle design, and to Dr. Feng Chen for providing the computing resources necessary to complete the experiments.

During the final preparation of this manuscript, we utilized the GPT-4 language model provided by OpenAI to assist in identifying and correcting typographical and grammatical errors. The use of this tool was restricted solely to the polishing stage and did not influence the study’s conceptual framework, research methodology, data analysis, or conclusions. All substantive content and intellectual contributions remain those of the authors, and the AI assistance served only to ensure clarity and precision in the final written presentation.
\bibliography{custom} 

\appendix
\onecolumn

\clearpage
\section{Appendix}
\subsection{Figures}
\label{Appendix Figures}

\begin{figure}[!hb]
    \centering
    \includegraphics[width=1\textwidth]{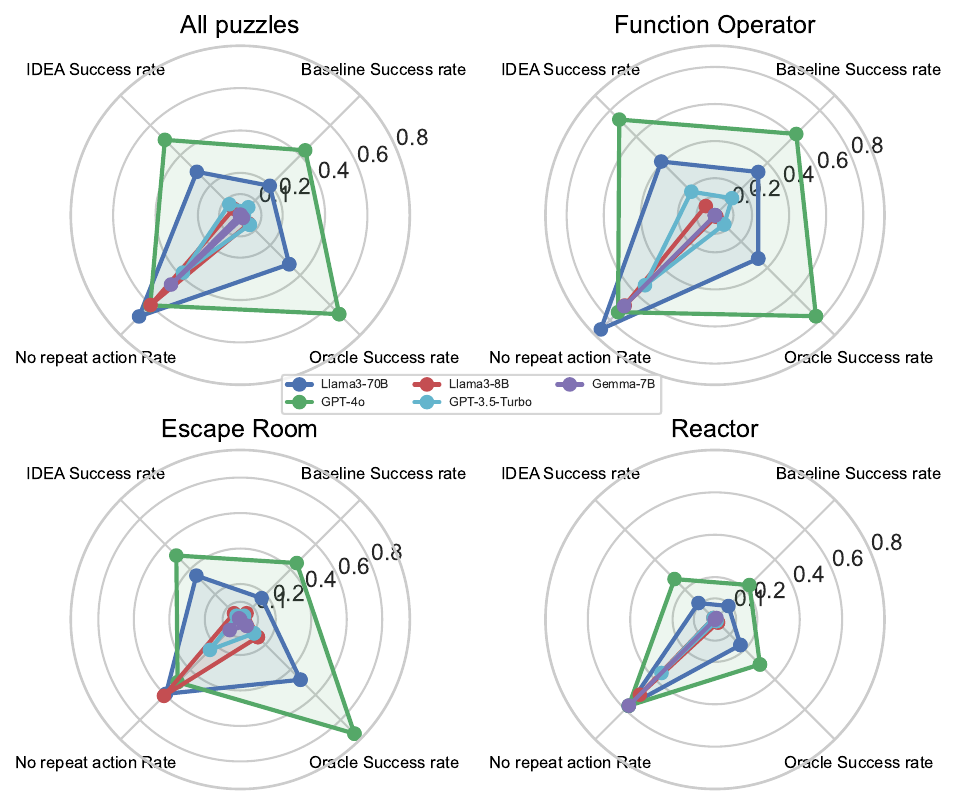}  
    \caption{\small The scaled performance radar plot shows varying performances across different puzzle types. GPT-4o leads, followed by Llama 70B, GPT-3.5, Llama 8B, and Gemma 7B.}
    \label{fig:radar_performace_plot}
\end{figure}

\begin{figure}[!htb]
    \centering
    \includegraphics[width=1\textwidth]{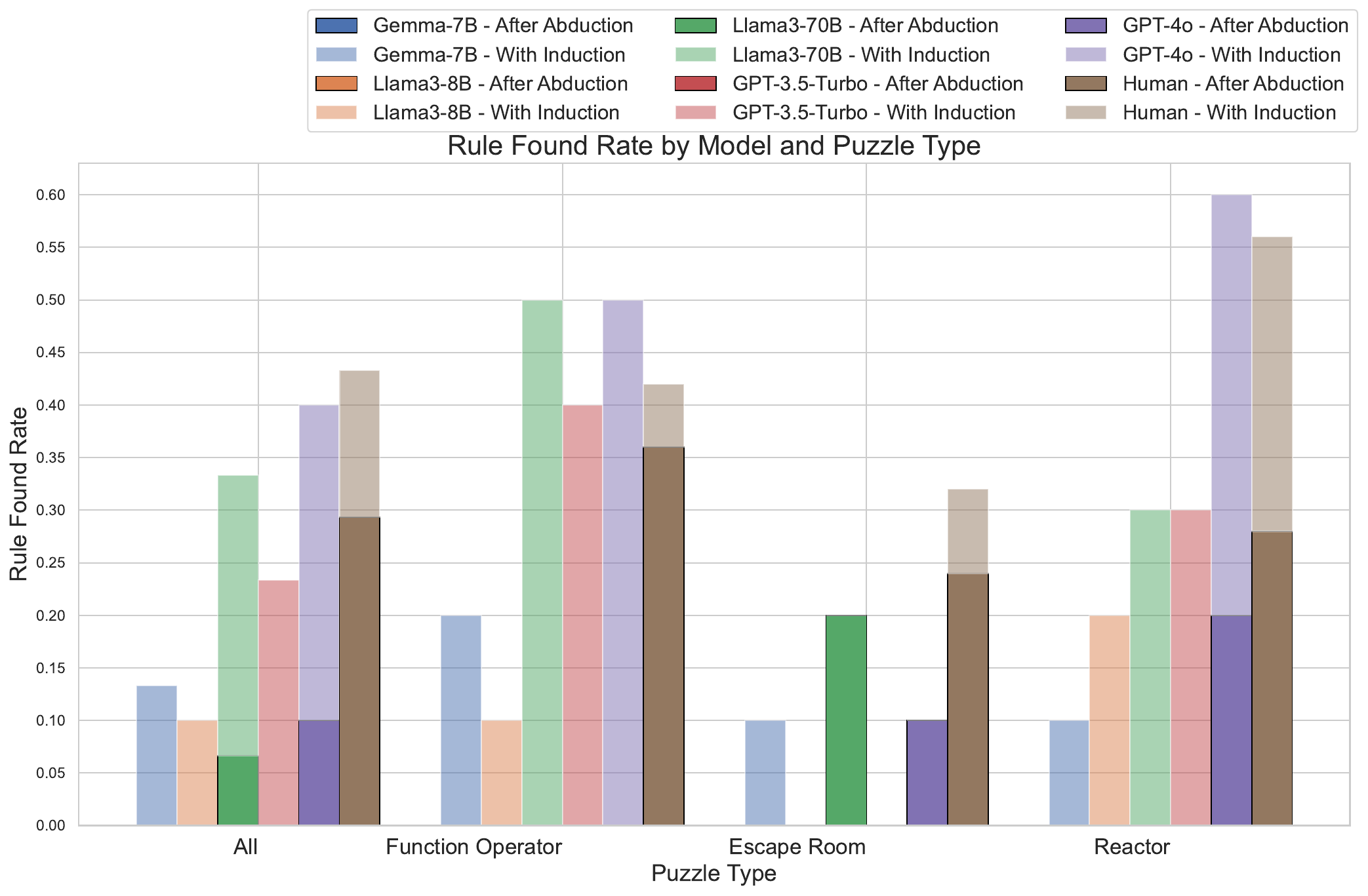}  
    \caption{\small Although agents continuously refine their hypotheses toward the ground truth rule, identifying the exact rule remains a challenging task. According to our evaluation, humans have a 43.3\% success rate in finding the ground truth rule, with 13\% of these discoveries occurring during the abduction stage and 30\% during the induction stage. In contrast, LLM agents exhibit a different pattern. They successfully identify the ground truth rule in approximately 30\% of puzzles, with nearly all of these discoveries occurring during the abduction stage and only 5\% achieved through interaction and induction. This highlights a significant limitation of current LLM agents, as they lack the ability to learn effectively from interactions. Consequently, the rule-learning patterns of LLM agents differ markedly from those of humans.}
    \label{fig:rule_found_rate_figure}
\end{figure}

\clearpage
\subsection{Tables}
\label{Appendix tables}

\begin{table}[!htb]
\centering
\caption{\small Function operator puzzle distribution}
\begin{tabular}{@{}ccccc@{}}
\toprule
\textbf{No. of Functions}  & \textbf{No. of Paramters} & \textbf{No. Terms} & \textbf{No. of Elementary Functions} & \textbf{No. of puzzles}\\ \midrule
 1 & 1 & 1 & 1 & 25 \\
 2 & 2 & 2 & 2 & 25 \\
 2 & 2 & 3 & 2 & 25 \\
 3 & 3 & 5 & 4 & 15 \\
 3 & 4 & 6 & 5 & 10 \\
\bottomrule
\end{tabular}
\label{tab:Function operator distribution}\\
\end{table}

\begin{table}[!htb]
\centering
\caption{\small Distribution of puzzles in the Escape Room scenario. For each number of paintings from 3 to 13, there are two visibility conditions: one where all paintings are initially visible, and one where the agent must take actions to reveal all paintings. Under each condition, there are five unique puzzles, resulting in a total of 10 puzzles per number of paintings.}
\begin{tabular}{ccc}
\toprule
\textbf{No. of Paintings} & \textbf{Visibility Condition} & \textbf{No. of Puzzles} \\ \midrule
3  & All visible at start            & 5 \\
3  & Requires actions to reveal      & 5 \\
4  & All visible at start            & 5 \\
4  & Requires actions to reveal      & 5 \\
\vdots & \vdots                      & \vdots \\
13 & All visible at start            & 5 \\
13 & Requires actions to reveal      & 5 \\
\bottomrule
\end{tabular}
\label{tab:Escape room distribution}
\end{table}

\begin{table}[!htb]
\centering
\caption{\small Distribution of Reactor puzzles across the four rule categories. Each category contains 25 puzzles, drawn from the same set of 25 distinct letter strings, which vary in length from 3 to 6 characters. Each puzzle requires the agent to synthesize a target string according to the specified rule.}
\begin{tabular}{@{}ccc@{}}
\toprule
\textbf{Reactor Rule} & \textbf{No. Initial Letters} & \textbf{No. of Puzzles} \\ \midrule
Simple Concatenation & 2, 3, 4, 5, 6 & 6, 8, 7, 2, 2 \\
Reverse Concatenation & 2, 3, 4, 5, 6 & 6, 8, 7, 2, 2 \\
Middle Insertion & 2, 3, 4, 5, 6 & 6, 8, 7, 2, 2 \\
Prefix Replacement & 2, 3, 4, 5, 6 & 6, 8, 7, 2, 2 \\
\bottomrule
\end{tabular}
\label{tab:Reactor distribution}\\
\end{table}

\begin{table}[!htb]
\centering
\caption{\small Average Number of Repeated Actions Per Puzzle: Repeating actions is a common pattern among LLM agents during rule-learning tasks. Even sophisticated models like GPT-4o often exhibit reduced action duplication when exploring environments using the \method agent. The implementation of this agent has been shown to mitigate this tendency across all models. However, Gemma-7B frequently generates nonsensical actions that are not recognized as duplicates. Consequently, a duplication rate of 0.02 does not necessarily indicate that Gemma-7B effectively avoids repeating historical actions.}
\scriptsize
\begin{tabular}{@{}llcccc@{}}
\toprule
Setup & Model & All Puzzles & Function Operator & Escape Room & Reactor Puzzles \\ \midrule
\multirow{5}{*}{Deduction Only} & Gemma-7B & 6.54& 4.05& 8.74& 6.83 \\
 & Llama-8B & 4.91& 3.39& 2.85& 8.49 \\
 & Llama-70B & 3.39& 2.47& 1.44& 6.25 \\ & GPT-3.5-Turbo & 8.06& 7.52& 6.27& 10.38 \\
 & GPT-4o & 2.51& 2.01& 0.65& 4.86 \\
\midrule
\multirow{5}{*}{Baseline} & Gemma-7B & 7.39& 6.12& 8.05& 8.01 \\
 & Llama-8B & 6.26& 6.41& 3.24& 9.13 \\
 & Llama-70B & 3.36& 1.25& 1.59& 7.23 \\
 & GPT-3.5-Turbo & 6.87& 6.85& 4.09& 9.66 \\
 & GPT-4o & 2.68& 1.86& 0.19& 5.99 \\
\midrule
\multirow{5}{*}{\method} & Gemma-7B & 5.0& 3.65& 7.1& 4.26 \\
 & Llama-8B & 3.77& 3.92& 2.73& 4.65 \\
 & Llama-70B & 1.73& 0.43& 0.72& 4.05 \\
 & GPT-3.5-Turbo & 5.67& 4.69& 3.55& 8.76 \\
 & GPT-4o & 2.37& 1.32& 1.25& 4.53 \\

\midrule
 & Human & 0.76& 0.46& 1.6& 0.22 \\
\bottomrule
\label{tab:repeat_count}
\end{tabular}
\end{table}

\subsection{\method agent detail}

\subsubsection{Environment Entities}
\label{appendix:Environment Entities}
\begin{itemize}
\item[$\bullet$] $\textbf{Agent}$: Represents the entity focused on rule-learning and problem-solving, comprising the following components:
    \begin{itemize}
        \item[*] \textbf{Goal} $(\mathbf{G})$: The objective of the agent, articulated in natural language.
        \item[*] \textbf{Buffer Memories} $(\mathbb{\Tilde{M}}:=\{\mathbf{\Tilde{m}_1,\Tilde{m}_2,\dots,\Tilde{m}_n}\})$: This temporary storage holds all newly generated information during the agent's exploration, including actions taken, outcomes of each action, and observations. This is where the most recent activities are initially recorded.
        \item[*] \textbf{Memories} $(\mathbb{M}:=\{\mathbf{m_1,m_2,\dots,m_n}\})$: This permanent memory stores all significant observations and facts from the beginning of the task. When the agent forms new assumptions and plans, the contents of the Buffer Memories are evaluated; non-essential details like are discarded, while important facts and observations are transferred to the permanent Memories. This architecture ensures that each time the agent revises its hypotheses, it can clearly distinguish which observations occurred under the new assumptions and plan.
        \item[*] \textbf{Hypotheses} $(\mathbf{H})$: The current hypotheses formulated by the agent to explain all the observations, are expressed in natural language.
        \item[*] \textbf{Plan} $(\mathbf{P})$: The current strategy devised by the agent to either test the correctness of the existing hypotheses or to leverage these hypotheses to achieve the goal, also represented in natural language.
        \item[*] \textbf{Action Space} $(\mathbb{A})$: A set of potential actions available to the agent, determined by its current hypotheses and plan. The Action Space is dynamic and can change in response to interactions with the environment. For example, after investigating a fridge, the agent gains the additional option to open the fridge and inspect its contents.
    \end{itemize}
\item[$\bullet$] $\textbf{Objects}$ $(\mathbb{O})$: Represents all interactive entities within the environment that provide the agent with a means to interact with the world. A single object in this set is denoted as $\mathbf{O}$.
    \begin{itemize}
        \item[*] \textbf{Description} $(\mathbf{D_o})$: A concise description of the object, detailing its nature and potential uses, presented in natural language.
        \item[*] \textbf{Predefined interactive actions} $(\mathbf{O_A}:=\{\mathbf{\tilde{a}_1,\tilde{a}_2,\dots,\tilde{a}_n}\})$: A set of actions that are predefined for each object. Each action is described in natural language, explaining its purpose. Additionally, each action is associated with a coded function that processes the agent's inputs and produces an effect, potentially altering the environment based on these inputs.
    \end{itemize}
\end{itemize}

\subsubsection{Interactive Functions}
\begin{itemize}
\item[$\bullet$] $\textbf{Perceptual Action}$:= $\mathbf{\hat{a}}(\mathbf{O})$: An action automatically added to the agent's action space for all objects within the same scope. Upon perceiving an object, the agent gains the ability to interact more detailedly with it, adding its interactive actions to the $\mathbf{S}$.
\item[$\bullet$] $\textbf{Interactive Action}$:= $\mathbf{\Tilde{a}(D_o,G,H,P,I}, \mathbb{\Tilde{M},M})$: A predefinec action that triggers a pre-coded effect based on the agent's input $\mathbf{I}$. For example, in using a reactor, the agent decides the materials and their order of addition, and the reactor processes these inputs based on pre-coded rules to synthesize new materials.
\item[$\bullet$] $\textbf{Abductive Action}$:= $\mathbf{\Bar{a}(G},\mathbb{\Tilde{M}})$: An action based on initial observations, allowing the agent to formulate the first hypotheses and generate a new plan.
\item[$\bullet$] $\textbf{Inductive Action}$:= $\mathbf{\dot{a}(G,H,P},\mathbb{\Tilde{M},M})$: An action based on the current observations, goals, prior hypotheses, and previous plans, allowing the agent to refine hypotheses and generate new plans.
\item[$\bullet$] $\textbf{Deductive Action}$ := $\mathbf{\ddot{a}(G,H,}\mathbb{A,M,\Tilde{M}} \mathbf{)}$: An action based on the current memories, hypothesis, and action space that generates a plan to either validate the current hypothesis or leverage it to solve problems.
\item[$\bullet$] $\textbf{Action select}$:=$F_a(\mathbf{G,H,P,}\mathbb{\Tilde{M},M,A})\rightarrow \mathbf{a}$: A function where the agent selects an action from the action space, considering all gathered information.
\end{itemize}

With the definitions and entities described above, we can formalize our interactive, rule-learning process. The sequence begins with the agent selecting an action from the available action space. The agent then decides on an input based on the selected action. Once the action is executed, the environment responds by providing feedback to the agent. The outcome of this action results in changes to $\mathbb{\Tilde{M}, M, S},\mathbf{H, P}$ and $\mathbb{O}$, making the environment dynamic as the exploration process progresses. These changes reflect the agent's interactions and adaptations to the evolving conditions within the environment.

\subsubsection{Pseudocode of interactive rule learning procedure}
\label{Pseudocode}
\begin{algorithm*}
\caption{Agent rule-learning procedure}
\begin{algorithmic}[1] 
\Procedure{RulelearningLoop}{}
    \State $\text{Initialize } \mathbb{O,A}, \mathbf{G}$
    \State $\mathbb{\Tilde{M}} \gets \text{Initial Memories}$
    \State $\mathbb{M} \gets []$
    \State $\textbf{H} \gets \mathbf{\Bar{a}}(\mathbf{G,\mathbb{\Tilde{M},A}})$
    \State $\textbf{P} \gets \mathbf{\ddot{a}(G,H,}\mathbb{A,M,\Tilde{M}} \mathbf{)}$
    \State $\mathbb{\Tilde{M}}.\text{add}(\text{``You established a new \textbf{H} and \textbf{P}.''})$
    \State $\text{\#step} \gets 0$ \Comment{Initialize step counter}

    \While{$\mathbf{G} \text{ not achieved} \text{ and } \text{step\_count} \leq \text{max\_step}$}
        \State $\mathbf{a} \gets F_a(\mathbf{G,H,P,}\mathbb{\Tilde{M},M,A})$ \Comment{Select an action from the action space}
        \If{$\mathbf{a}$ is a perceptual action}
            \State $\text{action\_result} \gets \text{execute\_perceptual\_action}(\mathbf{a}, \mathbf{O})$
            \State $\mathbb{A} \gets \text{update\_action\_space}(\text{action\_result})$
            \State $\mathbb{\Tilde{M}}.\text{add}(\text{action\_result})$ \Comment{Record result to buffer memory}
        \ElsIf{$\mathbf{a}$ is an interactive action}
            \State $\mathbf{I} \gets \text{decide\_input}(\mathbf{a}, \mathbf{D_o}, \mathbf{G}, \mathbb{\Tilde{M}, M}\mathbf{, H, P})$ \Comment{Agent decide Input for this action}
            \State $\text{action\_result} \gets \text{execute\_interactive\_action}(\mathbf{a, I})$
            \State $\mathbb{O} \gets \text{update\_states}(\text{action\_result})$ \Comment{update state of interactive objects}
            \State $\mathbb{A} \gets \text{update\_action\_space}(\text{action\_result})$ \Comment{Action may change action space}
            \State $\text{\#step} = \text{\#step} + 1$ \Comment{\textbf{Only interactive action increment step count}}
            \State $\mathbb{\Tilde{M}}.\text{add}(\text{action\_result})$ \Comment{Record result to buffer memory}
        \ElsIf{$\mathbf{a}$ is an inductive action}
            \State $\textbf{H} \gets \mathbf{\dot{a}(G,H,P},\mathbb{\Tilde{M},M})$
            \State $\textbf{P} \gets \mathbf{\ddot{a}(G,H,}\mathbb{A,\Tilde{M},M} \mathbf{)}$
            \State $\mathbb{M}.\text{filter\_add}(\mathbf{\Tilde{M}})$\Comment{Drop non-observational log and add the rest to $\mathbb{M}$}
            \State $\mathbb{\Tilde{M}} \gets []$\Comment{\text{Empty buffer memory}}
            \State $\mathbb{\Tilde{M}}.\text{add}
            (\text{``You established a new \textbf{H} and \textbf{P}.''})$
        \EndIf        
    \EndWhile
\EndProcedure
\end{algorithmic}
\end{algorithm*}
\clearpage

\subsection{Computational Budget}
\label{comp_budget}
For each setting (the Oracle agent, the Baseline, and the IDEA agent), we ran 300 different puzzles. On average, each API call’s input prompt is about 2,000 tokens, and the output is about 300 tokens. Completing each puzzle requires roughly 25 API calls (including both interaction and reasoning). Because each action changes the environment and updates the memory state, the input tokens cannot be cached, and each puzzle ends up using around 50,000 input tokens and 7,500 output tokens.
With 300 puzzles in a single setting, that totals approximately 15,000,000 input tokens and 2,250,000 output tokens per setting. Since we test three settings (Oracle, Baseline, and IDEA), a single model uses about 45,000,000 input tokens and 6,750,000 output tokens across all puzzles and settings.
For the open-source models, we used eight RTX-A6000 GPUs. LLaMA3-70B took about five days to complete its tasks, while LLaMA3-8B and Gemma-7B each required about 2.5 days. In total—across all three open-source models, three settings, and 300 puzzles—the open-source runs consumed roughly 80 GPU-days.
The experimental cost is significantly higher compared to traditional QA datasets, primarily because each puzzle requires about 25 API calls in sequence and the context grows rapidly during solving. In the future, we plan to develop methods to reduce costs by organizing the input more efficiently, leveraging cached output capabilities, and incorporating a filtering mechanism to manage long-term memory.

\subsection{Prompt example}
\label{Prompt example}
\subsubsection{Function Operator Puzzles}
\begin{figure}[!htb]
    \centering
    \includegraphics[width=\textwidth]{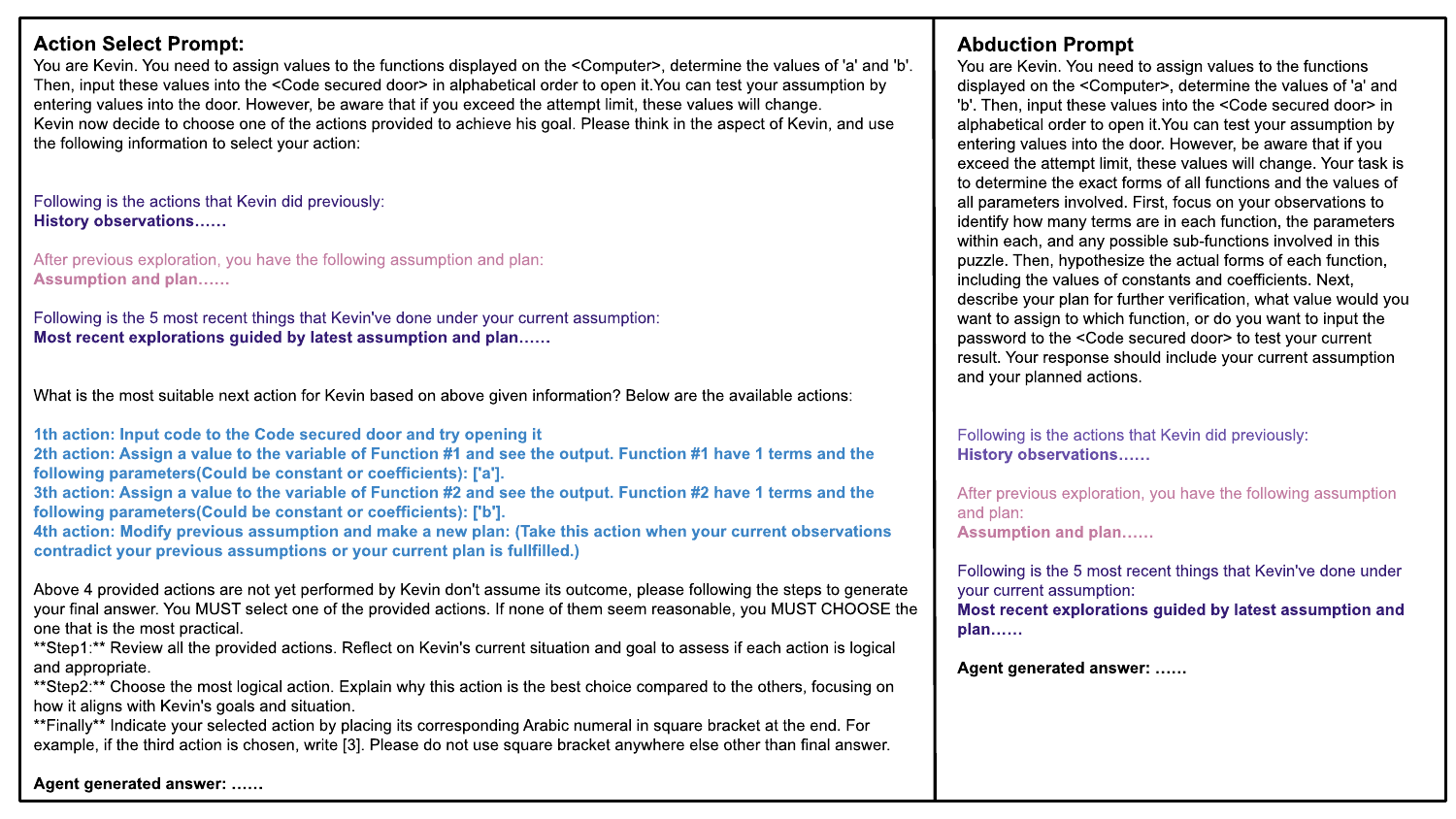}
    \caption{\small Prompt of Function Operator Puzzles, Action select and Deduction.}
    \label{fig:function operator prompt}
\end{figure}
\begin{figure}[!htb]
    \centering
    \includegraphics[width=\textwidth]{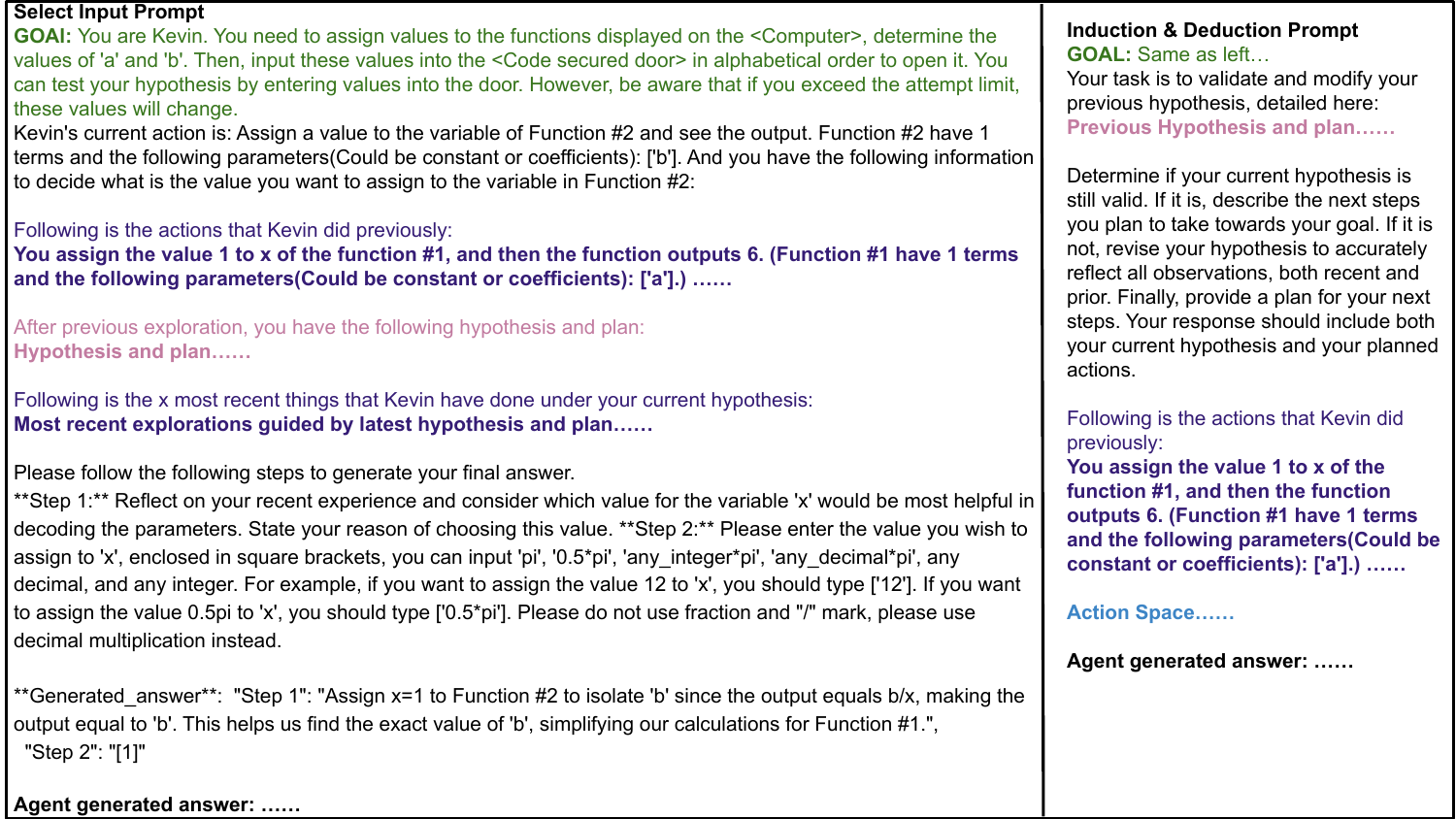}
    \caption{\small Prompt of Function Operator Puzzles, Interactive input and Induction.}
    \label{fig:function operator prompt 1}
\end{figure}

\begin{figure}[!htb]
    \centering
    \includegraphics[width=\textwidth]{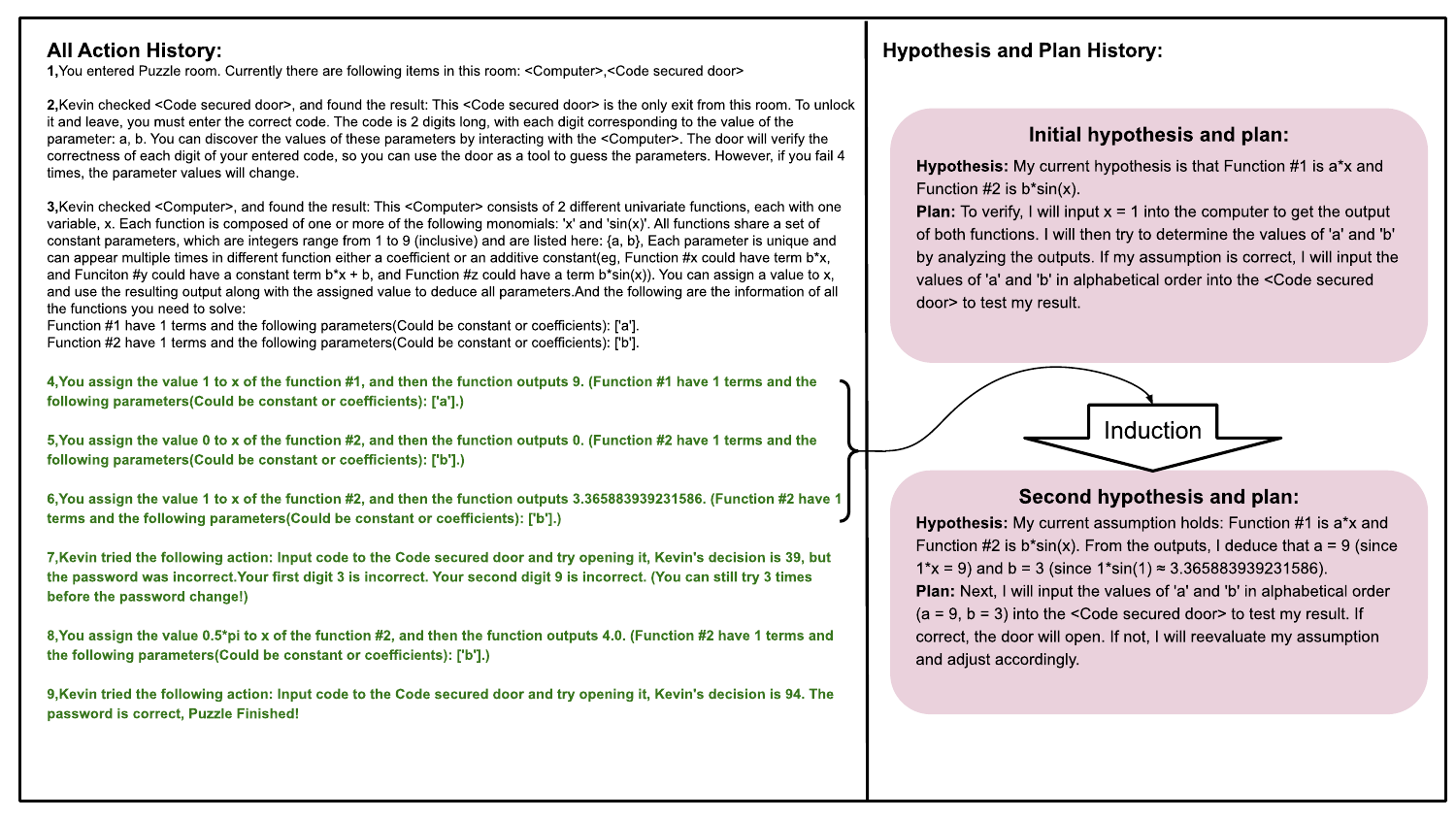}
    \caption{\small An example of Function Operator Puzzles is provided where actions marked in \textcolor{deepgreen_1}{green} are interactive actions, while the rest are perceptual actions through which the agent reads and perceives necessary environmental information.}
    \label{fig:function operator example}
\end{figure}

\begin{figure}[!htb]
    \centering
    \includegraphics[width=\textwidth]{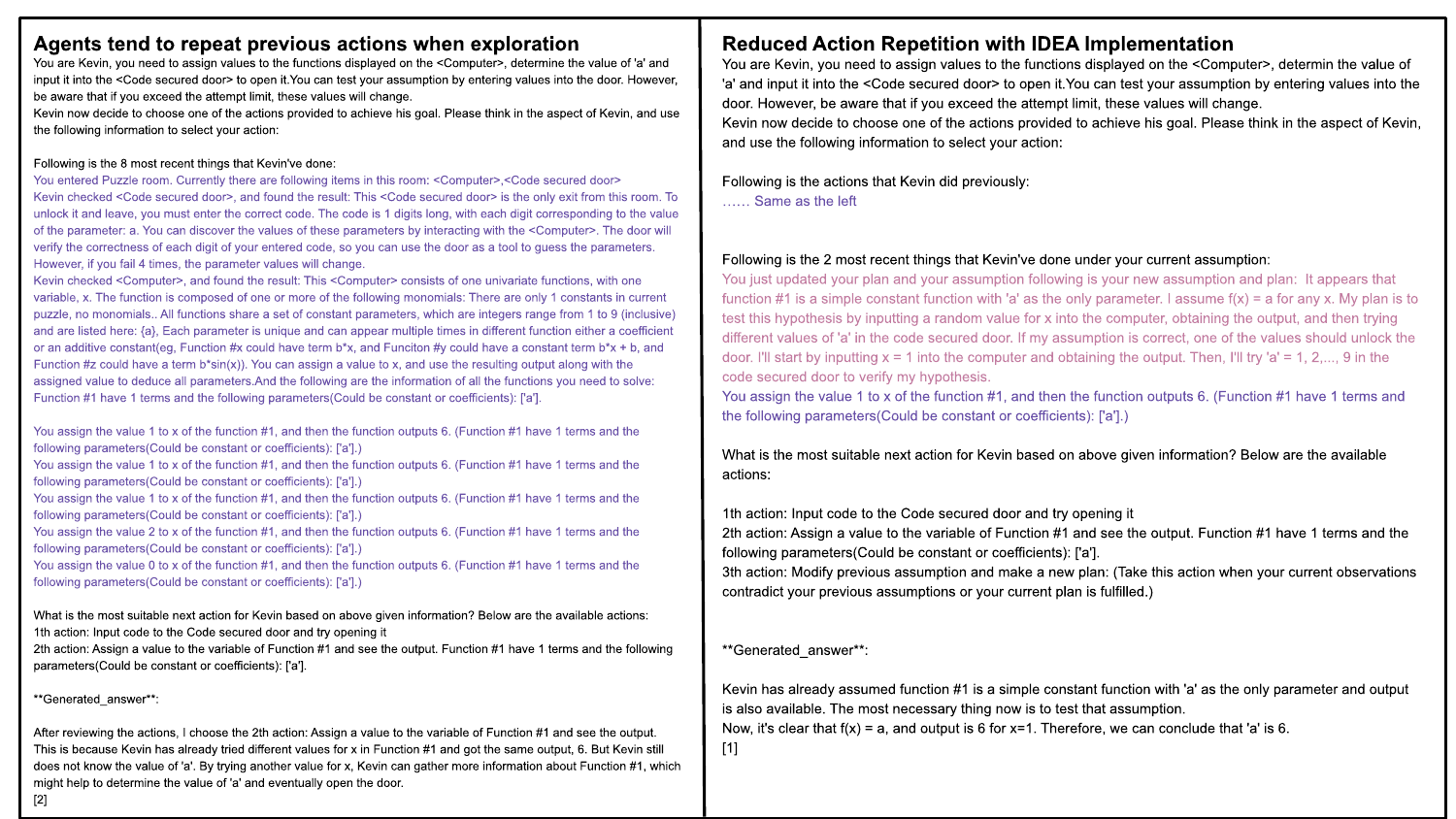}
    \caption{\small In the provided example, despite extensive exploration, the agent continues to assign multiple values to the function. In contrast, the \method agent hypothesizes that the function is simple, possessing only a single constant parameter. Consequently, this agent efficiently solves the puzzle by assigning just one value to the function.}
    \label{fig:repeatition example}
\end{figure}

\clearpage
\subsubsection{Escape Room Puzzles}
\begin{figure}[!htb]
    \centering
    \includegraphics[width=\textwidth]{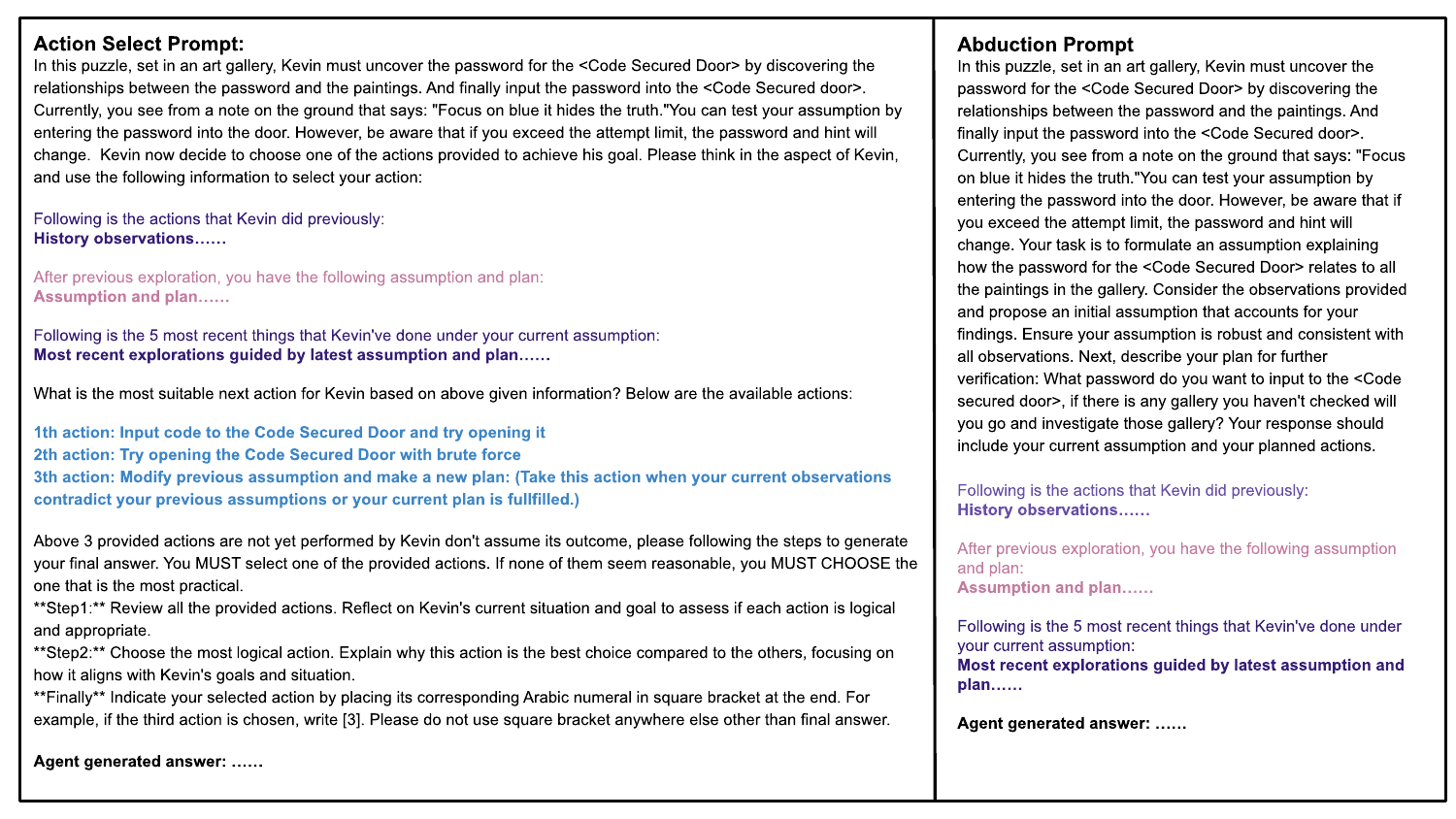}
    \caption{\small Prompt of Escape Room puzzles, Action select and Abduction.}
    \label{fig:art gallery prompt}
\end{figure}

\begin{figure}[!htb]
    \centering
    \includegraphics[width=\textwidth]{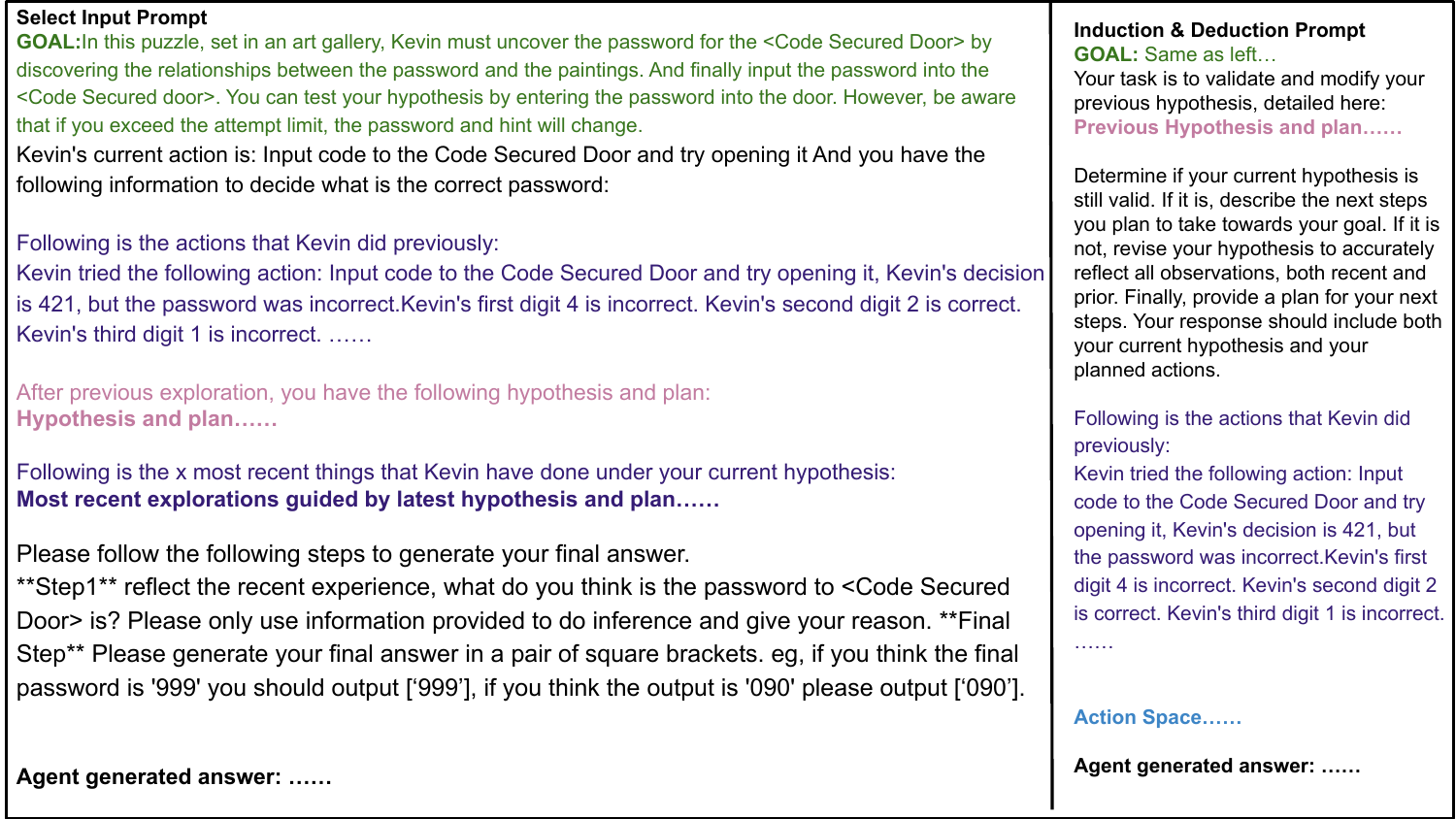}
    \caption{\small Prompt of Escape Room puzzles, Interactive Input and Induction.}
    \label{fig:escape_room_prompt_induction}
\end{figure}

\begin{figure}[!htb]
    \centering
    \includegraphics[width=\textwidth]{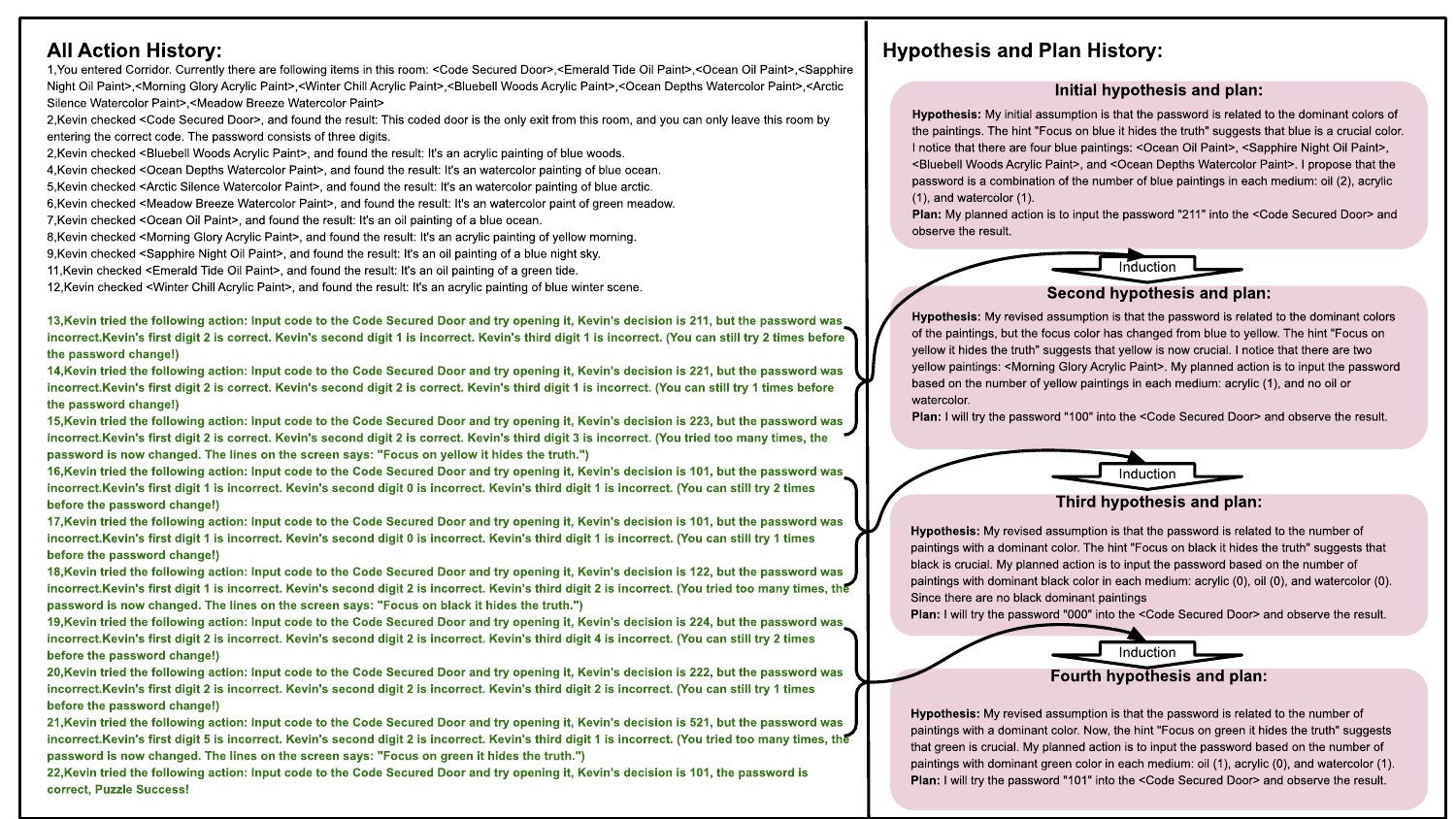}
    \caption{\small An example of Escape Room is provided where actions marked in \textcolor{deepgreen_1}{green} are interactive actions. The remaining actions, which are perceptual, allow the agent to read and gather necessary environmental information. In the given example, although the agent successfully guesses the correct rule behind the observations, it fails to adhere to its plan and assumptions. When inputting the password, the attempts do not align with the planned strategy, and it also makes repeated attempts (repeated 101 twice) that yield no useful results.}
    \label{fig:art gallery example}
\end{figure}

\clearpage
\subsubsection{Reactor Puzzles}
\begin{figure}[!htb]
    \centering
    \includegraphics[width=\textwidth]{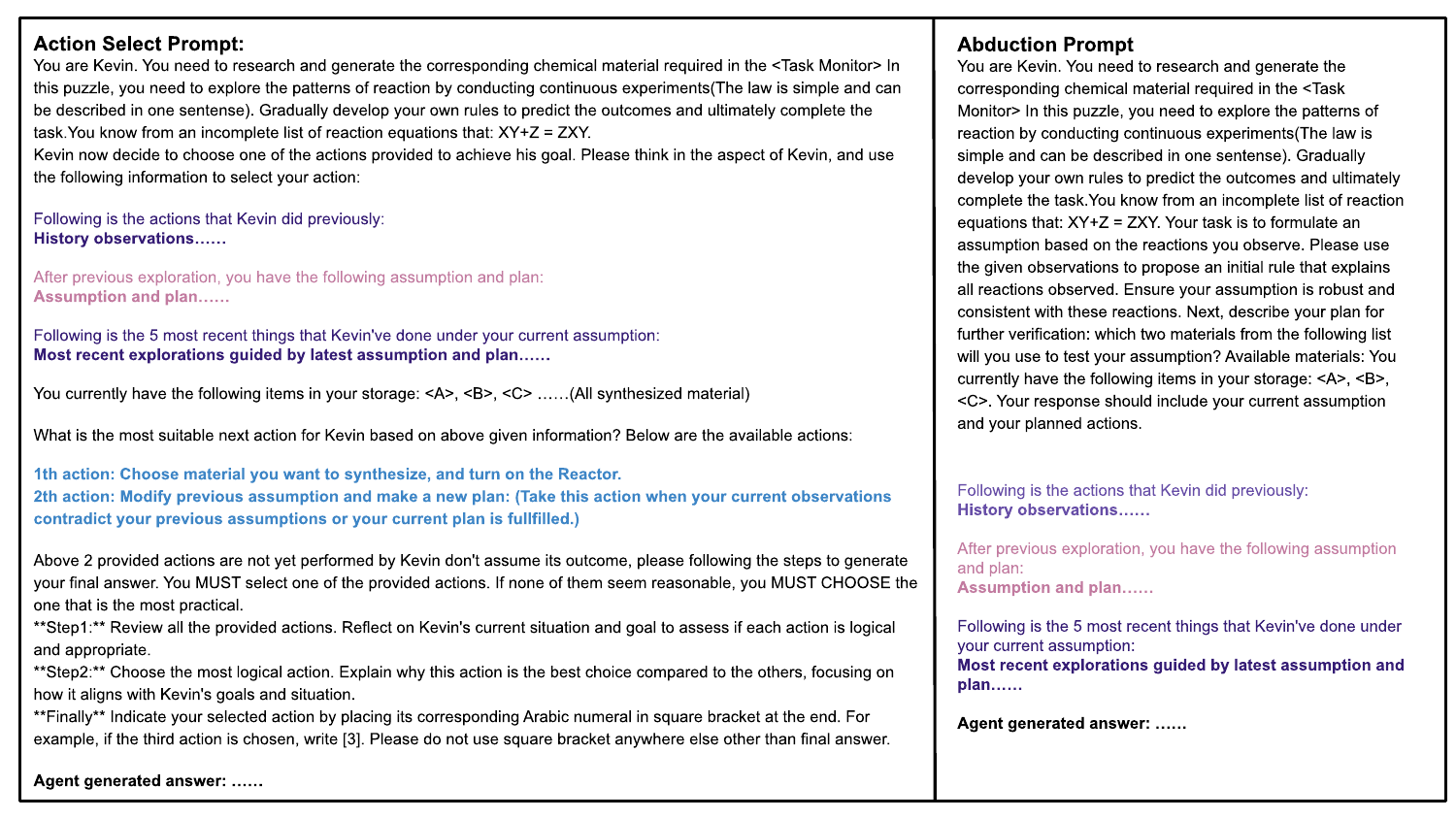}
    \caption{\small Prompt of Reactor Puzzles, Action select and Abduction}
    \label{fig:reactor prompt}
\end{figure}
\begin{figure}[!htb]
    \centering
    \includegraphics[width=\textwidth]{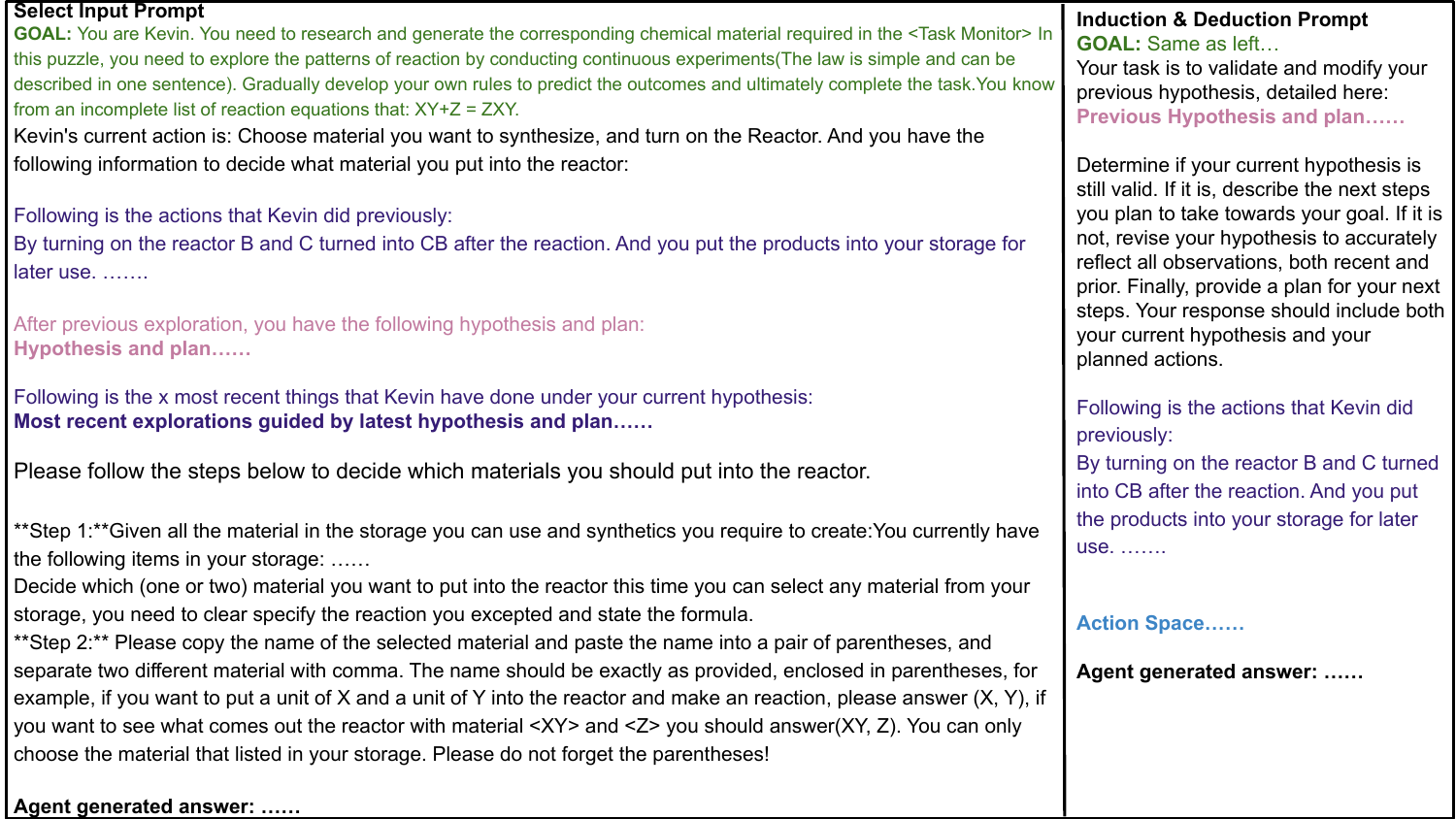}
    \caption{\small Prompt of Reactor puzzles, Interactive input and Induction.}
    \label{fig:reactor_prompt_induction}
\end{figure}

\begin{figure}[!htb]
    \centering
    \includegraphics[width=\textwidth]{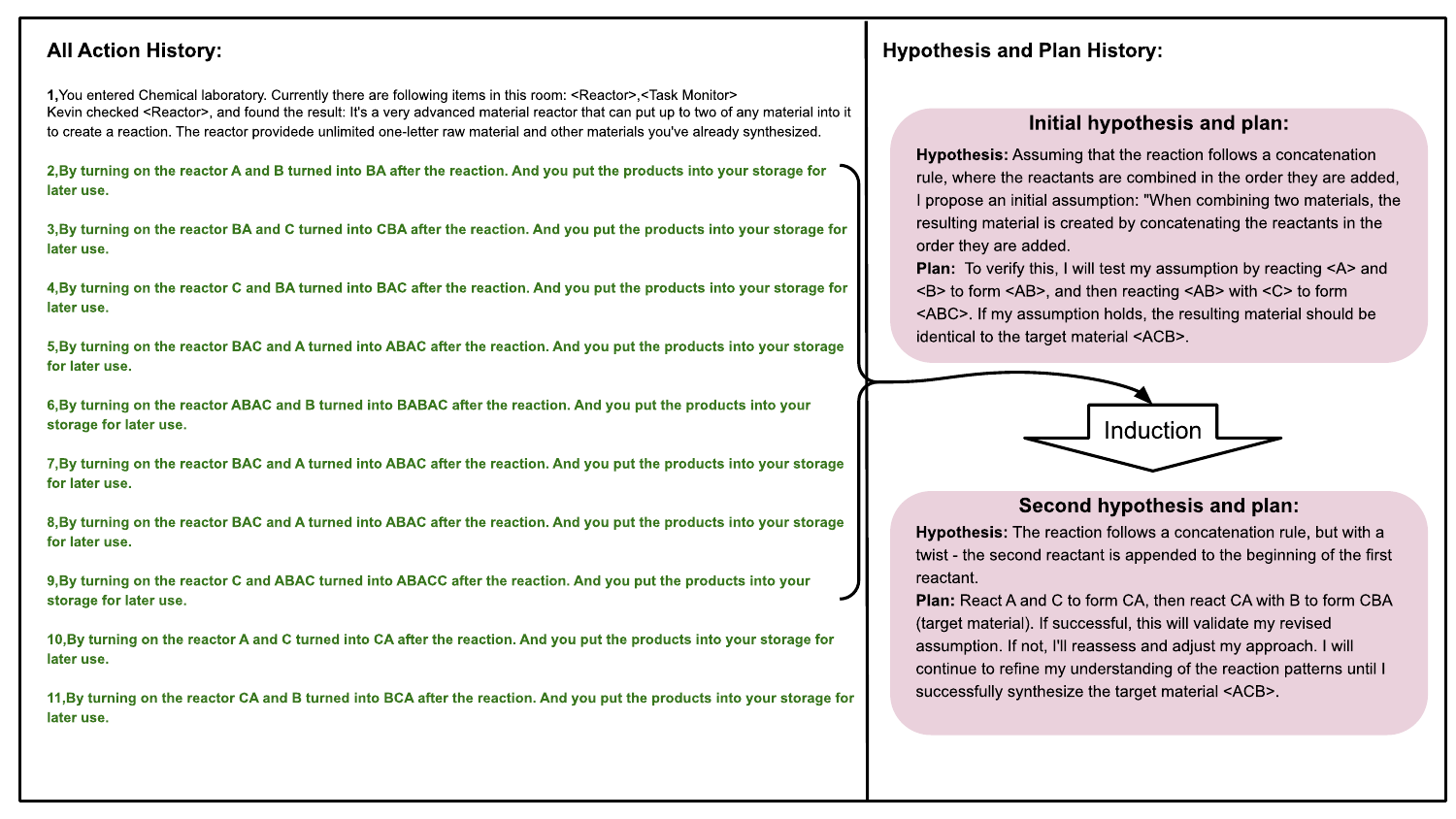}
    \caption{\small An example of Reactor Puzzles is provided where actions marked in \textcolor{deepgreen_1}{green} are interactive actions. The remaining actions are perceptual, allowing the agent to read and gather necessary environmental information. In the given example, the agent eventually realizes the flaws in its initial hypothesis and generates a correct one. However, the agent reaches the step limit before it can implement the solution, failing.}
    \label{fig:reactor example}
\end{figure}
\clearpage

\subsubsection{Task-Agnostic Prompts}
\label{Task-Agnostic Prompts}
\begin{tcolorbox}[title=Task-Agnostic Abduction Prompt , left=2mm,right=1mm,top=0mm, bottom=0mm,colback=white]
\begin{lstlisting}[style=plain]
f'''{GOAL} Your task is to develop general, clearly falsifiable, explanatory rules that explains your observations, a process known as abduction. Please consider the given observations and propose an initial hypothesis that explains them, make sure your hypothesis is robust and align with all your observations. Your response should include your current hypothesis and your planned actions.\nPlease keep your analysis, hypothesis, and planned actions as concise and precise as possible.'''

\end{lstlisting}
\end{tcolorbox}

\begin{tcolorbox}[title=Task-Agnostic Induction Prompt , left=2mm,right=1mm,top=0mm, bottom=0mm,colback=white]
\begin{lstlisting}[style=plain]
f'''{GOAL} Your task is to validate and modify your previous hypothesis, detailed here: {Previous_hypothesis}, using your new observations. Review your most recent observation: {buffer_memory_str}, to determine if your current hypothesis is still valid. If it is, describe the next steps you plan to take towards your goal. If it is not, revise your hypothesis to accurately reflect all observations, both recent and prior. Finally, provide a plan for your next steps. Your response should include both your current hypothesis and your planned actions. You can choose from the following actions:\n{action_space_str}\nPlease keep your analysis, hypothesis, and planned actions as concise and precise as possible.'''
\end{lstlisting}
\end{tcolorbox}

\subsection{Human participants interface}
\label{user interface section}
\begin{figure}[!htb]
    \centering
    \includegraphics[width=\textwidth]{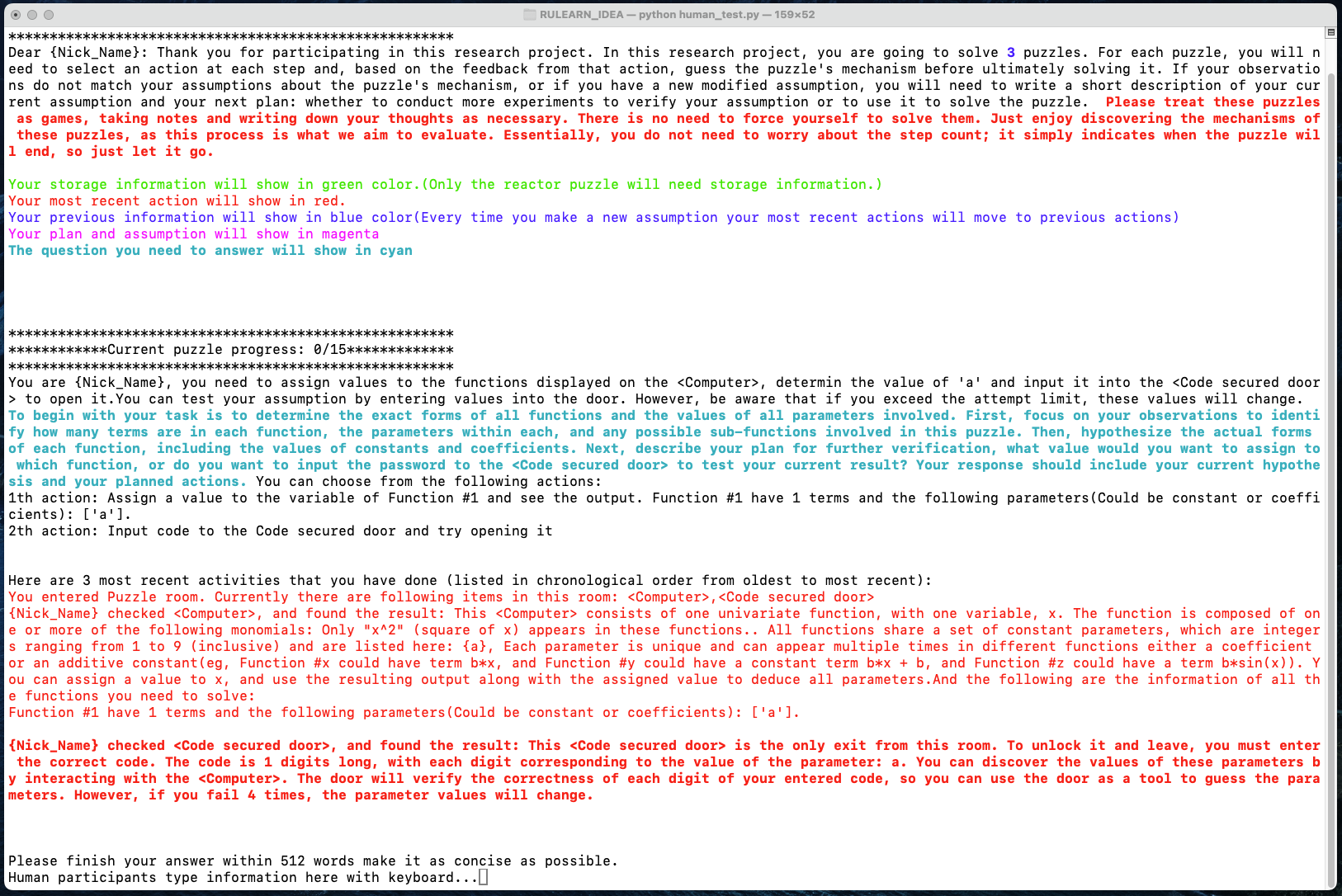}
    \caption{\small An example of what human participants see during the experiment. Human participants view the same prompt as the LLM agents, except for additional overview information clarifying their responsibilities in the project. The name ``\{{Nick\_name\}}'' is a pseudonym used to differentiate participants; we require participants to use only fictitious nicknames rather than their real names to ensure that the data collected is de-identified.}
    \label{fig:user interface}
\end{figure}

\clearpage
\subsubsection{Hallucination examples}
\label{Hullicination examples}
\begin{figure}[!htb]
    \centering
    \includegraphics[width=\textwidth]{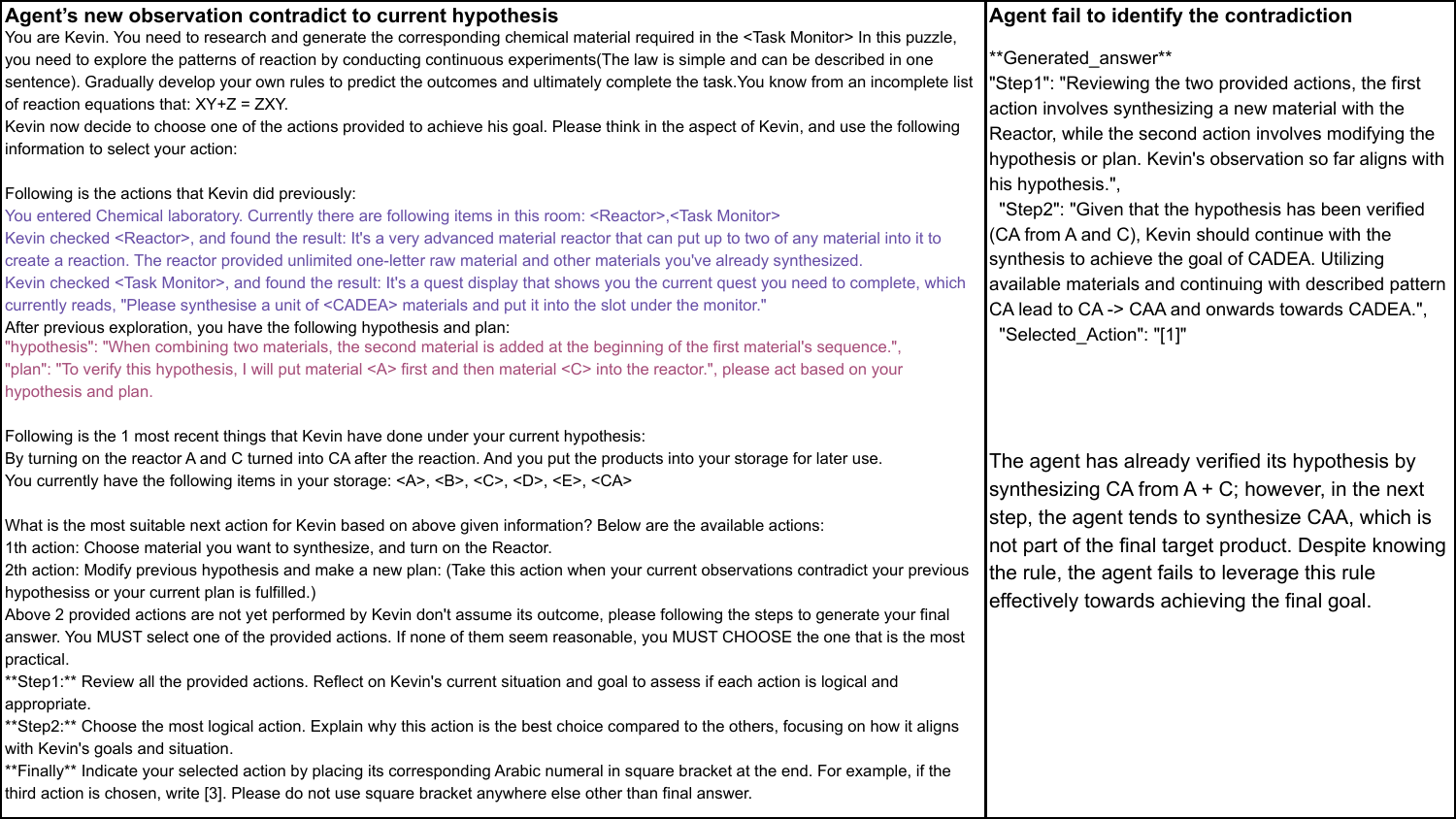}
    \caption{\small Example of GPT-4o agent failing to leverage rule towards the result.}
    \label{fig:fail_laverage_rule}
\end{figure}
\begin{figure}[!htb]
    \centering
    \includegraphics[width=\textwidth]{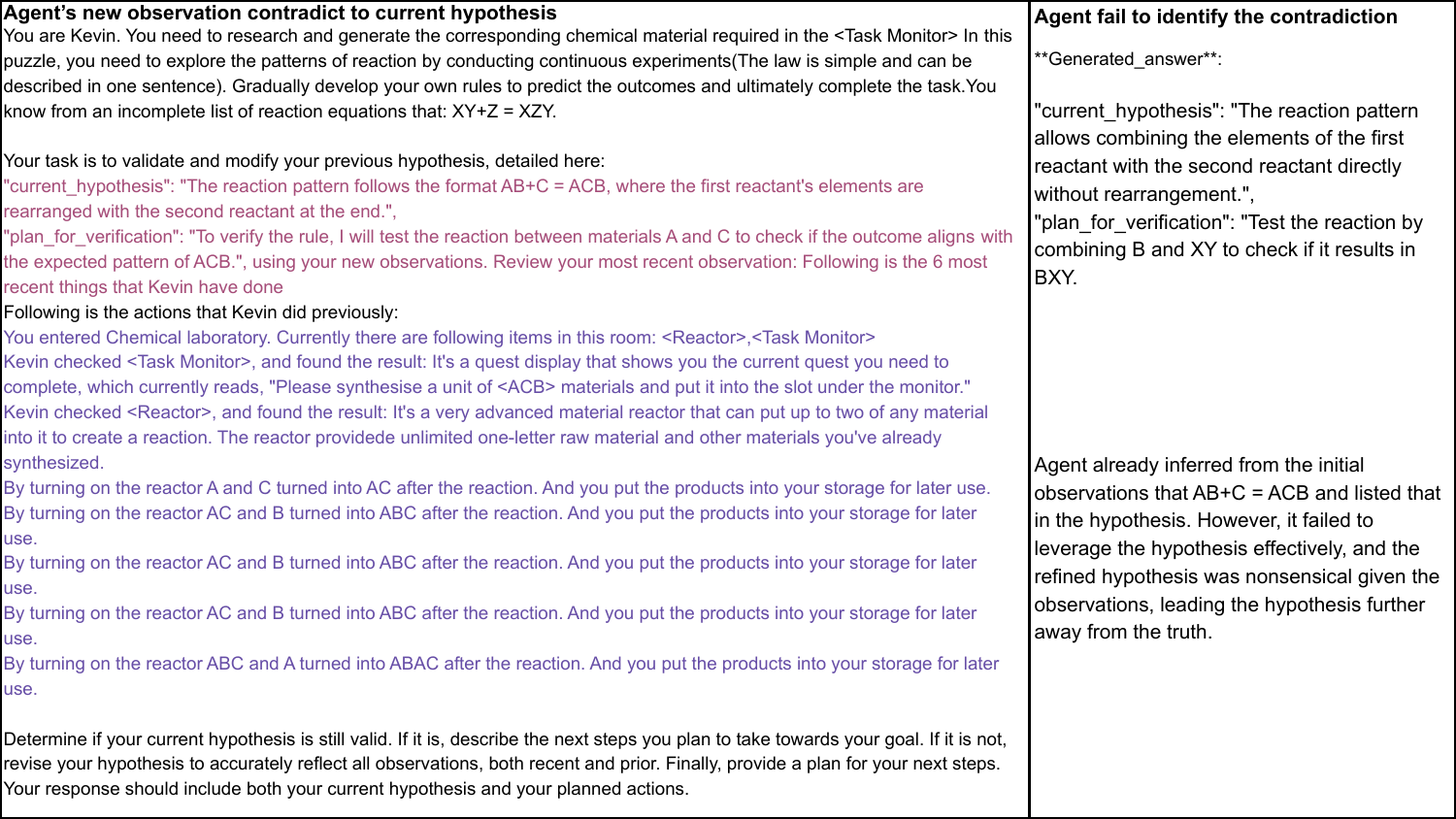}
    \caption{\small Example of hallucination by GPT-4o when refining previous hypothesis.}
    \label{fig:gpt-4o-hallucination_example}
\end{figure}
\begin{figure}[!htb]
    \centering
    \includegraphics[width=\textwidth]{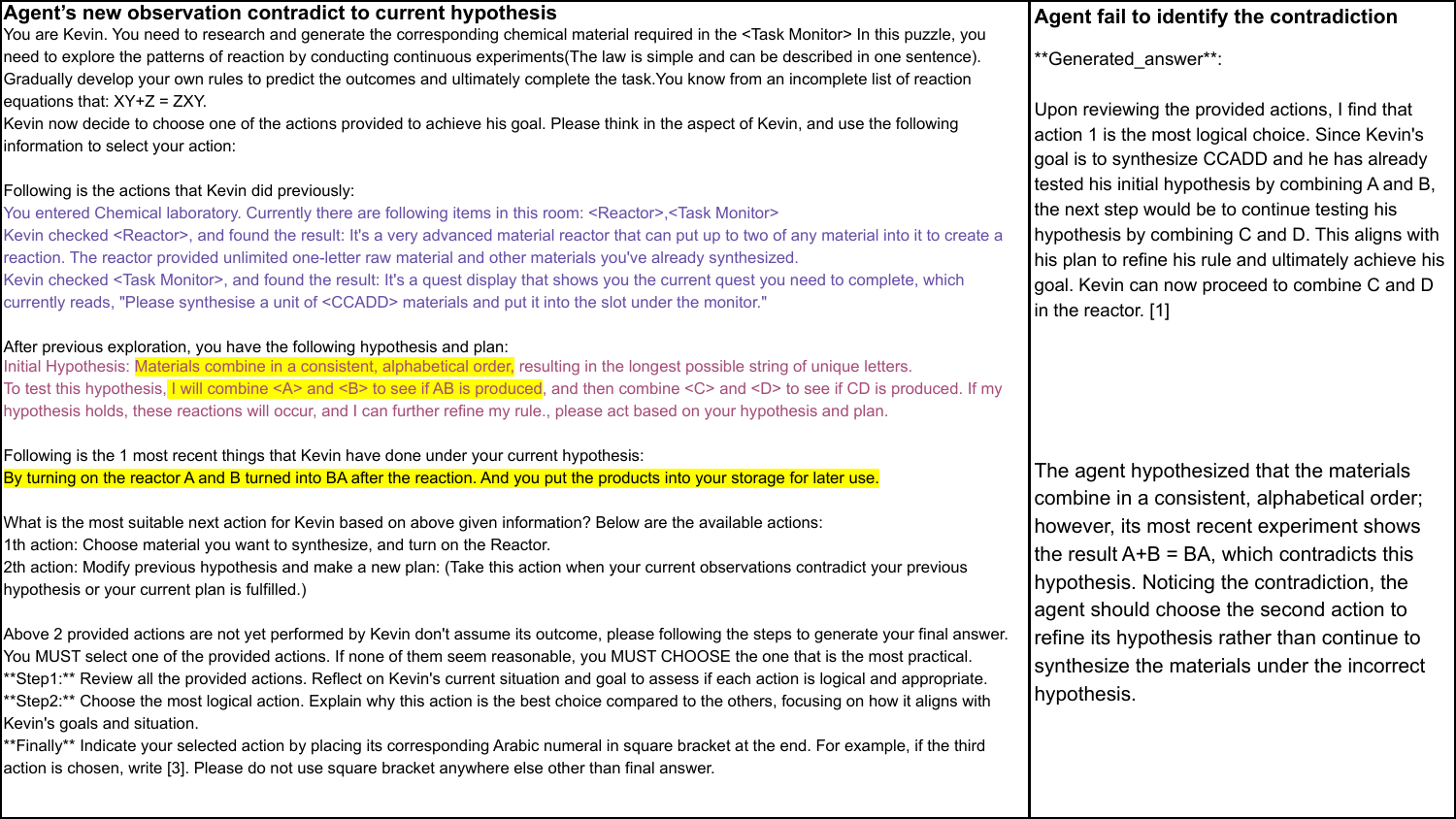}
    \caption{\small Example of Llama-3 70B agent failing to detect a contradiction in experimental results.}
    \label{fig:fail_notice_contradiction}
\end{figure}
\begin{figure}[!htb]
    \centering
    \includegraphics[width=\textwidth]{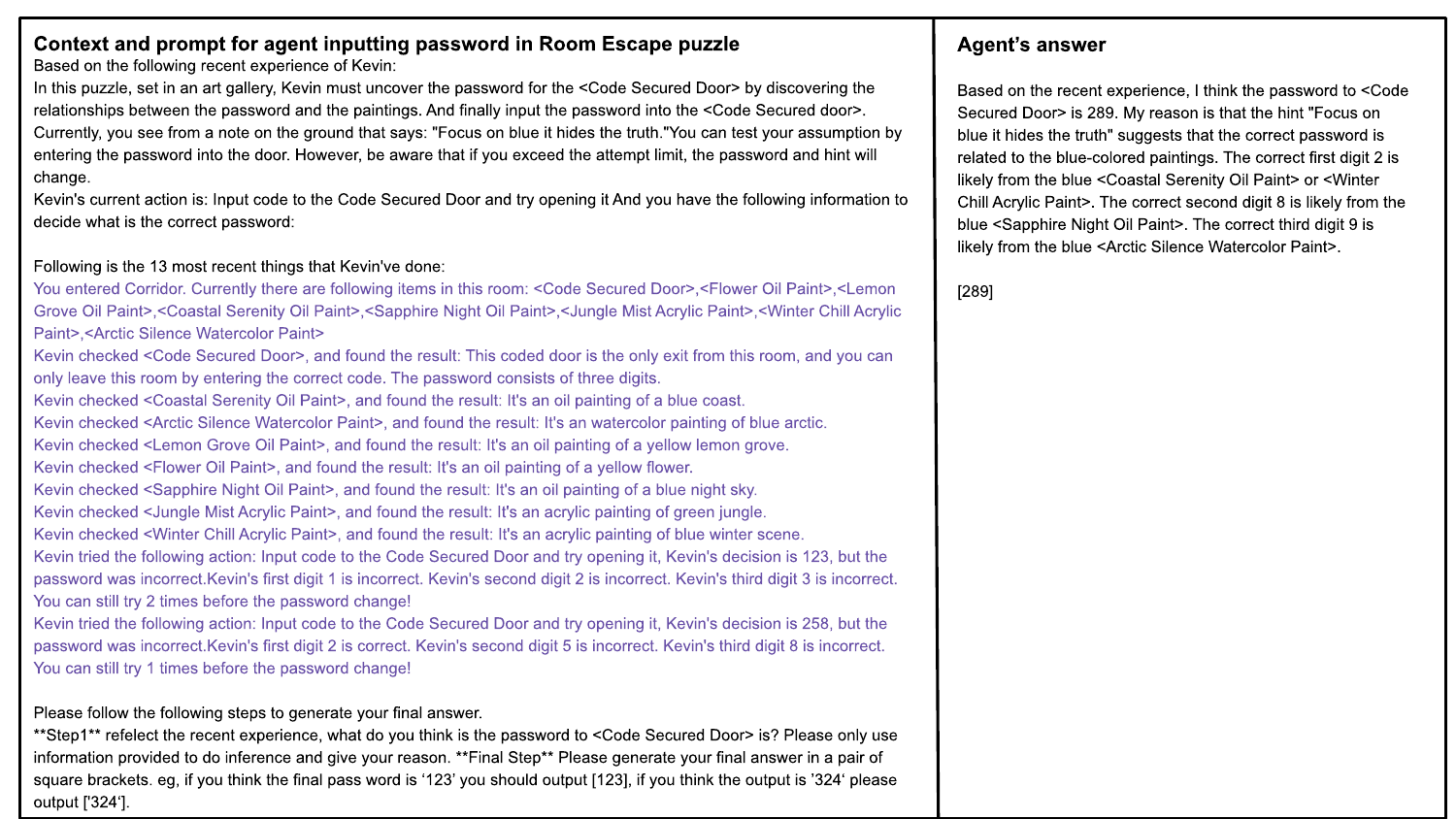}
    \caption{\small Example of hallucination by Llama-3 70B when generating a password.}
    \label{fig:art gallery hallucination}
\end{figure}

\clearpage
\subsection{Puzzle Examples}
\label{Puzzle Examples}
\begin{figure}[!htb]
    \centering
    \includegraphics[width=\textwidth]{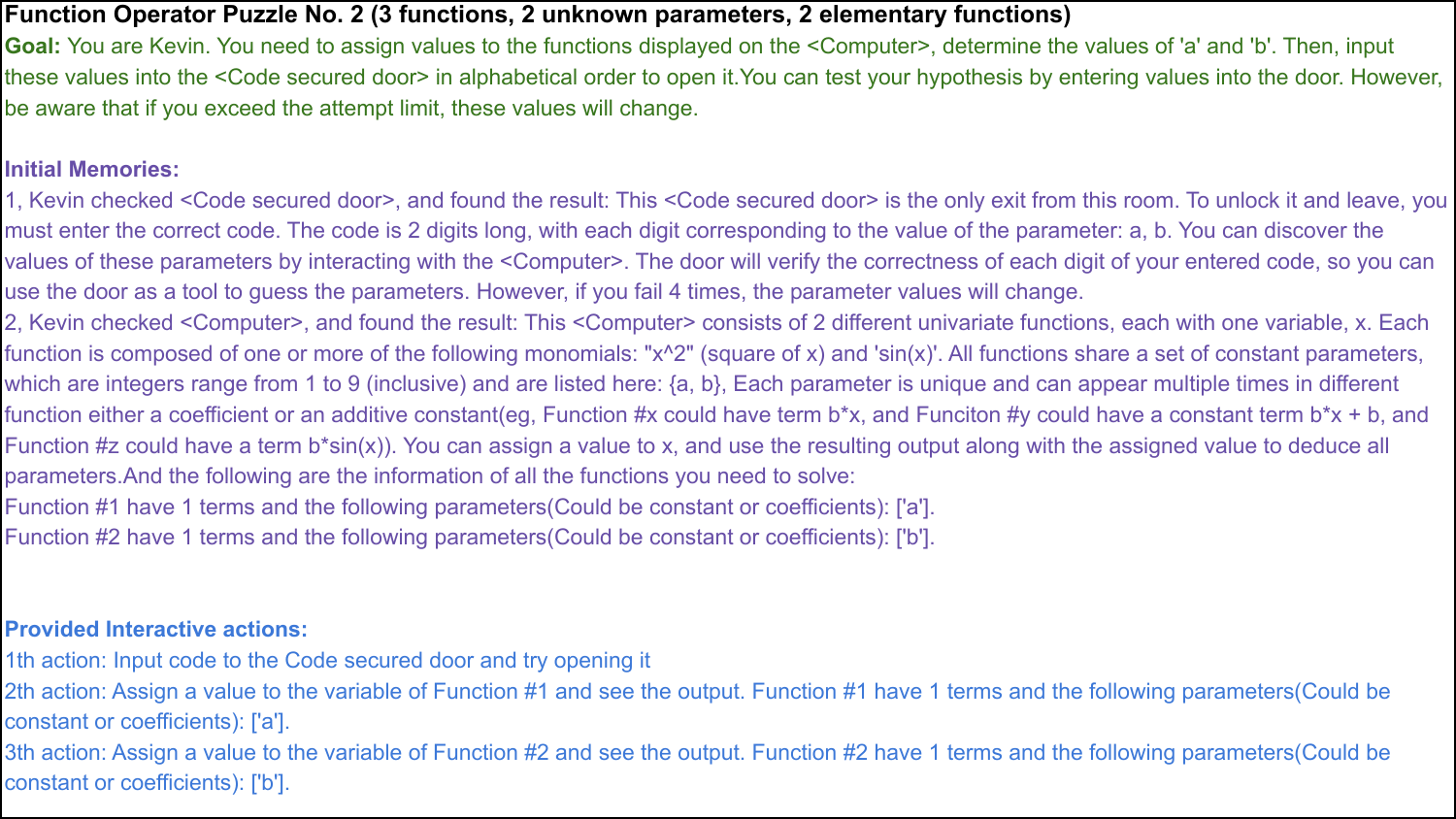}
    \caption{Function operator puzzle No. 2.}
    \label{fig:function_operator_setting_example_2}
\end{figure}
\begin{figure}[!htb]
    \centering
    \includegraphics[width=\textwidth]{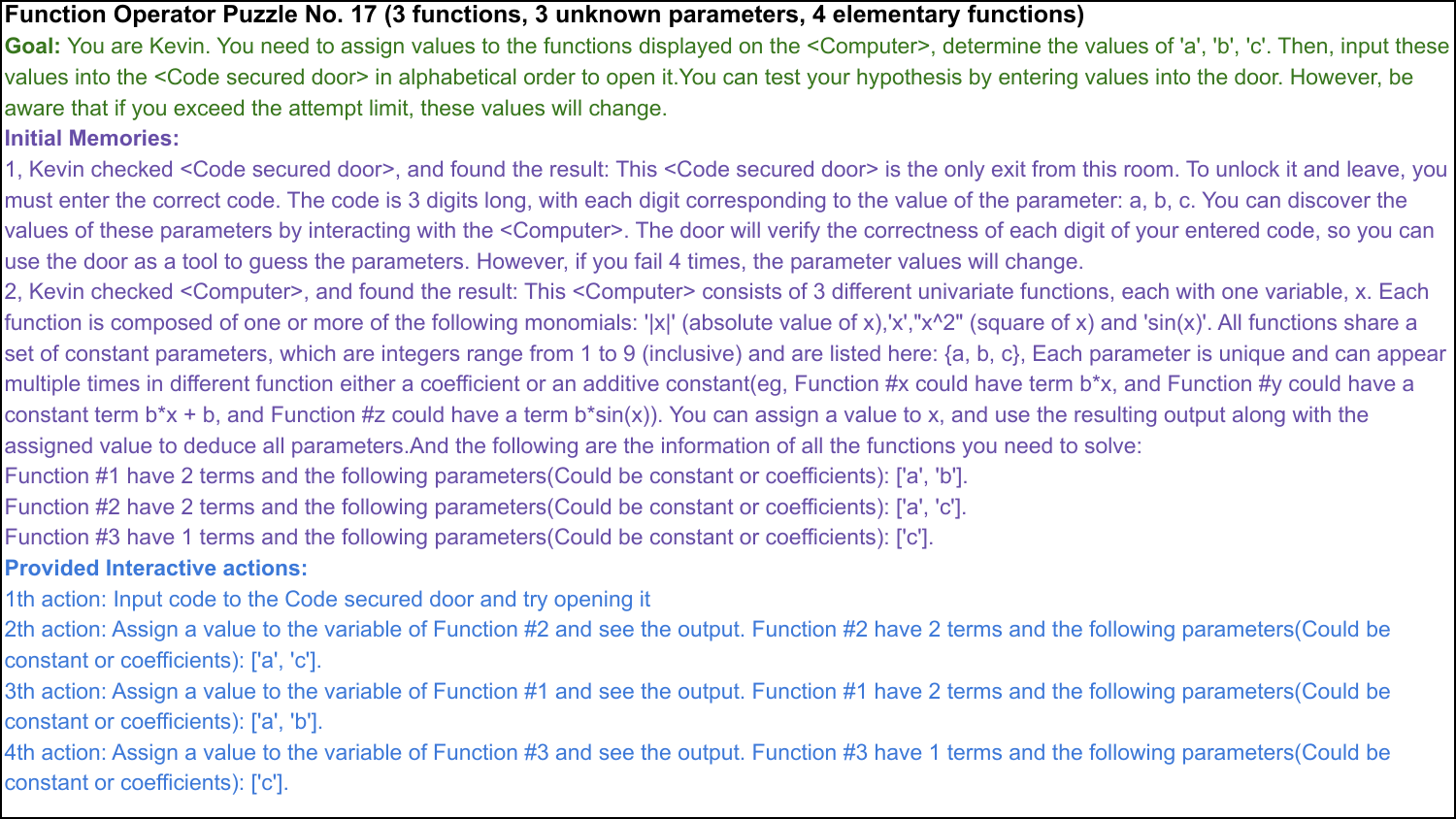}
    \caption{Function operator puzzle No. 17.}
    \label{fig:function_operator_setting_example_17}
\end{figure}
\begin{figure}[!htb]
    \centering
    \includegraphics[width=\textwidth]{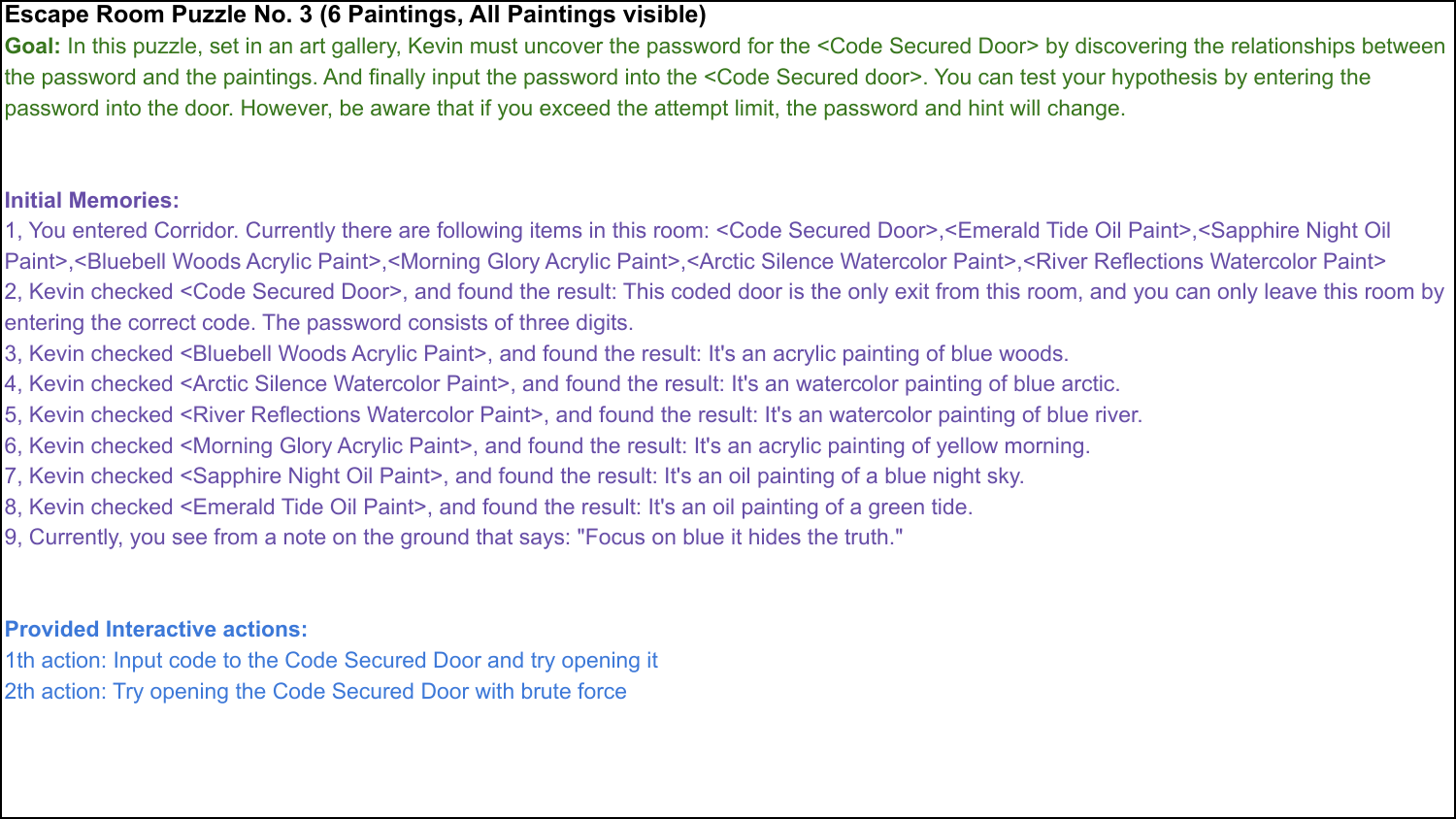}
    \caption{Escape room puzzle No. 3}
    \label{fig:escape_room_setting_example_1}
\end{figure}
\begin{figure}[!htb]
    \centering
    \includegraphics[width=\textwidth]{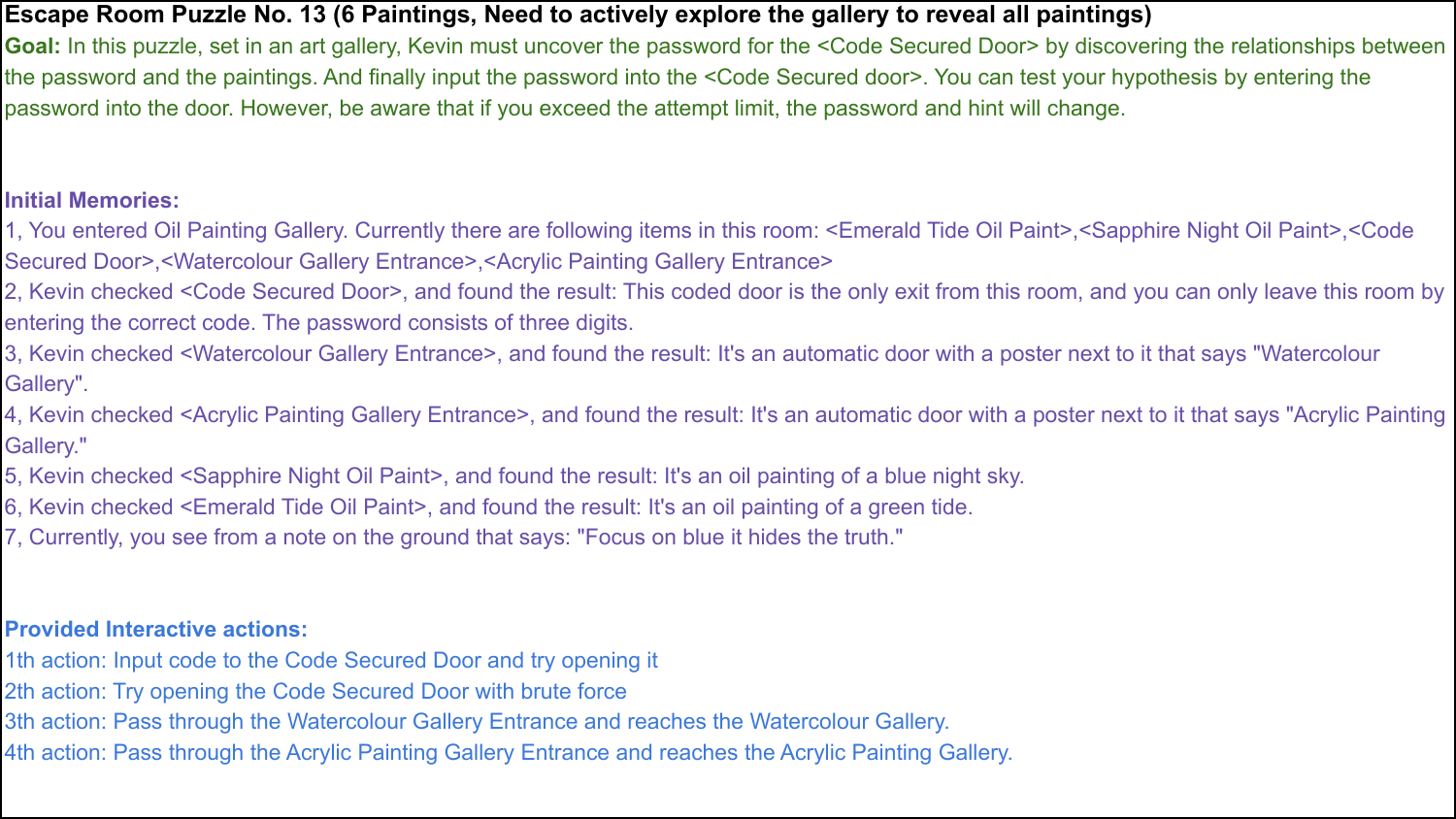}
    \caption{Escape room puzzle No. 13}
    \label{fig:escape_room_setting_example_2}
\end{figure}
\begin{figure}[!htb]
    \centering
    \includegraphics[width=\textwidth]{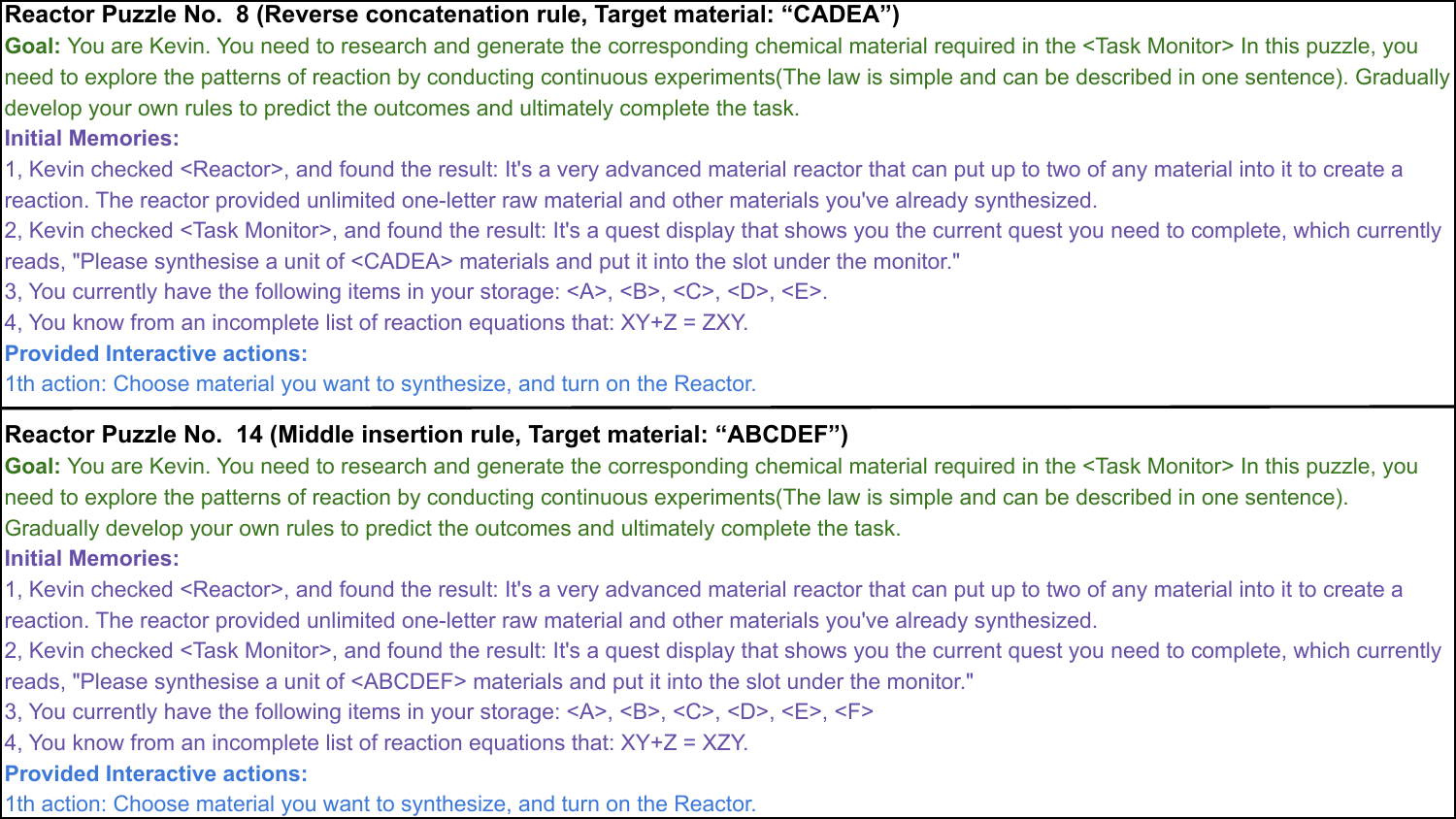}
    \caption{Reactor puzzle No. 8 and No. 14}
    \label{fig:reactor_setting_example_1_2}
\end{figure}

\end{document}